\documentclass{article}

\usepackage{microtype}
\usepackage{graphicx}
\usepackage{booktabs} 

\usepackage{hyperref}

\usepackage[accepted]{icml2023}

\usepackage{amsmath}
\usepackage{amssymb}
\usepackage{mathtools}
\usepackage{amsthm}

\usepackage[capitalize,noabbrev]{cleveref}

\usepackage{bbm, dsfont}
\usepackage{subcaption}
\usepackage{tikz}
\usepackage{booktabs,rotating,multirow}
\usepackage{setspace}

\theoremstyle{plain}
\newtheorem{theorem}{Theorem} 
\newtheorem{observation}{Observation} 
\newtheorem{lemma}[theorem]{Lemma}
\newtheorem{corollary}{Corollary} 
\theoremstyle{definition}

\theoremstyle{remark}
\newtheorem*{remark}{Remark}

\icmltitlerunning{Causal Strategic Classification: A Tale of Two Shifts}

\renewcommand{\paragraph}[1]{\textbf{#1}\,\,}

\newcommand{\red}[1]{} 
\newcommand{\blue}[1]{} 
\newcommand{\green}[1]{} 
\newcommand{\purple}[1]{} 
\newcommand{\orange}[1]{} 

\newcommand\todo[1]{{\red{TODO: {#1}}}}

\newcommand\tocitec[1]{\red{[CITE: {#1}]}}
\newcommand\toref{\red{[REF]}}

\newcommand\extended[1]{{\orange{EXTENDED: {#1}}}}
\newcommand{\nir}[1]{\green{[NIR: {#1}]}}

\newcommand{\guy}[1]{\purple{[GUY: {#1}]}}

\newcommand{\naive}{na\"{\i}ve}

\newcommand\expect[2]{\mathbbm{E}_{#1}{\left[ {#2} \right]}}

\newcommand{\ind}[1]{\mathds{1}{\{{#1}\}}}

\DeclareMathOperator*{\sign}{sign}
\DeclareMathOperator*{\argmax}{argmax}
\DeclareMathOperator*{\argmin}{argmin}
\renewcommand*{\d}{\mathop{}\!\mathrm{d}}
\newcommand{\minus}{\text{--}}

\newcommand{\R}{\mathbb{R}}

\newcommand{\yhat}{{\hat{y}}}
\newcommand{\phat}{{\hat{p}}}
\newcommand{\qhat}{{\hat{q}}}
\newcommand{\hhat}{{\hat{h}}}
\newcommand{\xtilde}{{\tilde{x}}}
\newcommand{\xbar}{{\bar{x}}}
\newcommand{\ytilde}{{\tilde{y}}}

\newcommand{\wbar}{{\bar{w}}}

\newcommand{\loss}{{\ell}}

\newcommand{\shinge}{\loss_{\text{s-hinge}}}

\newcommand{\method}[1]{{\fontfamily{lmtt}\selectfont{{#1}}}}
\newcommand{\CSERM}{{\method{CSERM}}}
\newcommand{\CSERMreg}[1]{{\method{CSERM$_{\text{{#1}}}$}}}
\newcommand{\CSERMlambda}{\CSERMreg{$\lambda$}}
\newcommand{\CSERMoracle}{{\method{oracle}}}
\newcommand{\SERM}{{\method{SERM}}}

\newcommand{\ERM}{{\method{ERM}}}

\newcommand{\RRM}{{\method{RRM}}}

\newcommand{\RRMlet}{{\method{RRM$_{\le t}$}}}
\newcommand{\RRMc}{{\method{RRMc}}}

\newcommand{\bench}{{\method{ns-bench}}}
\newcommand{\CSERMWRONG}{{\CSERM$_{\textsc{wrong}}$}}
\newcommand{\CSERMDROPS}{{\CSERM$_{\textsc{discard}}$}}

\newcommand{\myindent}[1]{
\makebox[#1cm]{}
}
\begin{document}

\twocolumn[
\icmltitle{Causal Strategic Classification: A Tale of Two Shifts}



\icmlsetsymbol{equal}{*}

\begin{icmlauthorlist}
\icmlauthor{Guy Horowitz}{technion}
\icmlauthor{Nir Rosenfeld}{technion}
\end{icmlauthorlist}

\icmlaffiliation{technion}{Technion -- Israel Institute of Technology, Haifa, Israel}

\icmlcorrespondingauthor{Nir Rosenfeld}{nirr@cs.technion.ac.il}

\icmlkeywords{Machine Learning, ICML}

\vskip 0.3in
]



\printAffiliationsAndNotice{}  

\begin{abstract}

When users can benefit from certain predictive outcomes,
they may be prone to act to achieve those outcome,
e.g., by strategically modifying their features.
The goal in strategic classification is therefore to train predictive models that are robust to such behavior.
However, the conventional framework assumes that changing features does not change actual outcomes,
which depicts users as `gaming' the system.
Here we remove this assumption,
and study learning in a causal strategic setting where true outcomes do change.
Focusing on accuracy as our primary objective,
we show how strategic behavior and causal effects
underlie two complementing forms of distribution shift.
We characterize these shifts,
and propose a learning algorithm that balances between these two forces and over time,
and permits end-to-end training.
Experiments on synthetic and semi-synthetic data demonstrate the utility of our approach. 
\end{abstract}

\section{Introduction}

The field of strategic classification 
\citep{bruckner2012static,hardt2016strategic} 
studies learning in a setting where users can strategically respond to a learned classifier
by modifying their features, at some cost, to obtain favorable predictive outcomes.
Such behavior can be expected  
when predictions are used to inform decisions about users, 
and from which users stand to gain (or lose);
common examples include loans approval, university admissions, and job hiring.
The framework of strategic classification
succinctly captures a widespread form of tension that naturally arises
between a classifier and the users it targets,
and which applies broadly.
This has made it the target of much recent interest
\citep{dong2018strategic,miller2020strategic,tsirtsis2020decisions,jagadeesan2021alternative,ghalme2021strategic,zrnic2021leads,levanon2021strategic,levanon2022generalized,estornell2021unfairness,lechner2021learning,ahmadi2022classification,nair2022strategic,eilat2022strategic,barsotti2022transparency}.

As a learning problem,
strategic classification is appealing 
in that its simple and clean formulation permits
and feasible practical challenges. 
But simplicity comes at a price,
and the framework's general applicability is hindered by its reliance on a set of strong assumptions. 
As part of a growing community effort to extend strategic classification beyond its original narrow form,
our goal in this work is to take one step towards making strategic classification more flexible.
%
In particular, here we target one of the key assumptions in strategic classification, which is the assumption that true outcomes $y$ do \emph{not} change when features $x$ are modified.
Under this assumption, strategic behavior amounts to gaming,
and users are depicted as acting to `fool' the classifier.
But in reality, this assumption rarely holds,
since actions taken by users to change predictions
can also change true outcomes.\looseness=-1

The observation that changes in $x$ can causally affect $y$
has been made by several authors \citep{miller2020strategic,shavit2020causal,bechavod2021gaming,harris2022strategic}.
But to date, works that have addressed this point focus primarily on the question of \emph{improvement}, i.e., whether (and how)
learning can incentivize users to change $x$ in ways that improve outcomes $y$.
While this is an important goal,
here we argue that the current perspective conflates
(i) the mere fact that $y$ can change, with
(ii) the desire for $y$ to change favorably.
But from a purely predictive point of view,
\emph{any} changes to $y$---whether for better or for worse---may deteriorate performance.
Hence, and given that the implications of causal strategic behavior on learning are not yet well-understood,
here we choose to focus entirely on the conventional goal of optimizing predictive accuracy,
and study appropriate notions of robustness.
We view this as an essential first step, intended to set the ground for more elaborate learning tasks such as incentivizing for improvement.\looseness=-1

\extended{\todo{running example?}}



Towards this, and aiming to remain as true as possible to the original formulation,
we seek to take the minimal necessary step beyond strategic classification
for introducing meaningful causal relations.
Our proposed setting,
which generalizes vanilla strategic classification,
can be described succinctly by a simple causal graph depicting the relations between different variable types: \emph{causal}, \emph{non-causal} (or `correlative'), and \emph{unobserved}.
The graph's structure defines the learning objective,
which in turn determines the precise form in which learning must be strategically-robust.
This formulation reveals where and how causality can impede learning,
and hints at how these challenges can be addressed.


In essence, strategic behavior can be viewed as entailing a certain form of distribution shift---with the key property that \emph{how} the distribution shifts depends on the choice of classifier, indirectly through how it shapes user responses
\citep{drusvyatskiy2022stochastic,maheshwari2022zeroth}.
Our first contribution is characterizing the role causality plays in this process.
When causality is absent, strategic updates $x{\mapsto}x'$
change $p(x)$,
but also $p(y|x)$; this is since the induced $p'(y|x')$ must 
`remember' the original $y$.
Conversely, 
we show that in a fully causal setting, 
learning reduces to a particular instance of \emph{decision-dependent covariate shift},
in which strategic behavior affects only the marginal $p(x)$;
In other words, causality `cancels out' the strategic effect on $p(y|x)$.
Thus,
the challenge in learning lies in correctly balancing between
two distinct notions of robustness.

Based on these insights,
our second contribution is a learning algorithm for strategic causal classification.
We focus on the setting where
users respond rationally and under a predetermined feature partition;
this places emphasis on coping with the uncertainty in $y$ introduced by the causal structure.
Here the challenge is that learning must simultaneously account for
(i) \emph{strategic} changes in $x$, in response to the learned classifier $f$; and
(ii) \emph{causal} changes in $y$, which result from changes in $x$.
The key to effective learning therefore lies in correctly decoupling strategic and causal effects;
towards this, and relying on our theoretical analysis,
our algorithm makes use of an estimated marginal density model $\phat(x)$, which is novel in this space.
As we show, our approach effectively separates informational uncertainty, which is irreducible,
from statistical uncertainty---which our approach efficiently reduces by making use of additional strategically-modified (i.e., `dirty') data.

Our approach becomes especially effective over \emph{time}:
here we make connections to the literature on \emph{performative prediction} \citep{perdomo2020performative},
and study causal strategic learning in a temporal setting and under retraining dynamics.
In standard strategic classification, 
learning is known to converge after a single time-step \citep{hardt2016strategic};
but this notion breaks once causal effects are introduced.
The fact that both $p(y|x)$ and $p(x)$ can now temporally change
poses a challenge, but also an opportunity:
using an appropriate form of regularization,
we show how learning can be made to incentivize feature updates that
reveal labeled information from under-represented areas of $p(x)$,
which contribute to an improved estimation of $p(y|x)$.

Finally, we conduct a series of experiments
that empirically validate our approach.
First, using synthetic data, we design experiments 
aimed at showcasing the challenges, pitfalls, and opportunities that arise when learning in causal strategic environments.
Then, we use real data (augmented with simulated responses)
to compare our approach to several baselines.
We report both quantitative and qualitative results,
and perform sensitivity analysis regarding to our structural assumptions.
Overall, our results shed light on the importance 
of accounting for causal effects in strategic setting,
and the need to correctly balance between these forces. 
All code is made publicly available and can be found at:\\
\url{https://github.com/guyhorowitz/CSC}.

\extended{\\
- intentionally do not do causal inference - want to stay within purely predictive \\
- $f$, $p$ \emph{induced} distribution $p^f$ \\
- list of contributions? if space...
}

\subsection{Related work}
\paragraph{Strategic classification.}
Since its introduction \citep{bruckner2012static,hardt2016strategic},
the literature on strategic classification has been growing rapidly.
Efficient learning algorithms have been proposed for the original batch setting
\citep{levanon2021strategic,levanon2022generalized},
as well as online formulations \citep{chen2020learning,ahmadi2021strategic}.
On the theoretical front,
\citet{zhang2021incentive,sundaram2021pac} extend PAC theory
via strategic VC analysis.
Ongoing efforts aim to extend the original setting 
to handle utilities that are unknown \citep{dong2018strategic},
noisy \citep{jagadeesan2021alternative},
estimated \citep{ghalme2021strategic, bechavod2022information, barsotti2022transparency},
allow for arbitrary preferences \citep{levanon2022generalized},
or are linked by a graph \citep{eilat2022strategic}.
Other works break or relax some core assumptions,
such as the order of play \citep{nair2022strategic}
or the role of time \citep{zrnic2021leads}.
Our work joins these efforts, with the aim of allowing true outcomes to causally change when features are modified.




\paragraph{Causal strategic learning.}
Several works blend causality with strategic learning,
but these focus almost exclusively on improvement.
\citet{kleinberg2020classifiers} study the problem of incentivizing agents to improve,
and 
\citet{alon2020multiagent,haghtalab2020maximizing} generalize their setting to multiple agents;
however, neither of these works
directly consider learning.
\citet{miller2020strategic} show that learning to incentivize 
improvement 
inevitably requires solving a non-trivial causal inference problem; thus,
coping with causality
requires making some assumptions about the underlying causal structure.
Some works 
make assumptions that permit causal inference, 
acting either through indirect experimentation in online learning
\citep{bechavod2021gaming}
or by using the published classifiers as instruments in an offline setting \citep{harris2022strategic};
these works, however, are restricted to regression.
Other works 
consider particular causal relations,
such as \citet{mendler2022predicting} who study predictions as interventions,
or \citet{chen2021linear} who decouple gaming and improving effects in both learning and evaluation.
Closest to ours is \citet{shavit2020causal}, who provide learning algorithms for improvement, estimation, and (to some extent) accuracy; however, they focus on linear regression (in which strategic responses are invertible), consider a realizable (linear) setting, and make assumptions that permit causal discovery.
Our work studies (agnostic) classification and
focuses predominantly on accuracy. \looseness=-1

\extended{
\paragraph{Performative prediction.}
\begin{itemize}
    \item PP studies the dynamics of learning in settings where deploying a classifier changes the distribution
    \item SC is often given as a special case, but is degenerate since it converges after one time step (when users best-respond)
    \item as we show, relaxing the '$y$ is unchanged' assumption introduces a temporal dimension, making it a `real' PP problem
    \item the benefit of casting causal SC as PP is that it gives (micro) structure -- this differs from most current works on PP that focus on giving global guarantees under sufficient (macro) properties
    \item most of the current works on PP talk about dynamics where both $x$ and $y$ are changing; however, they talk about abstract distribution maps and study macro-level properties whereas we inject micro-level structure by applying minimal change from SC and adding causal aspects.
\end{itemize}
}
\section{Problem setup}
We start by briefly describing standard strategic classification, and continue with our proposal for injecting causality.

\subsection{Standard strategic classification}
In the original formulation of strategic classification \cite{hardt2016strategic}, users have feature representations $x\in \mathcal{X}=\mathbb{R}^d$ and binary labels $y\in \mathcal{Y}= \left\{\pm1\right\}$. 
Let $p(x, y)$ be a joint distribution over (nonstrategic) features and labels. The primary goal in learning is to find a classifier $f : \mathcal{X}  \rightarrow \mathcal{Y}$ from a class $F$ that achieves high expected accuracy, 
given a train set
$S = \left\{(x_i, y_i)\right\}_{i=1}^m$ with $(x_i,y_i) \overset{iid}{\sim} p(x,y)$.
At test time, however, $f$ is evaluated on strategically-modified data, where users update features via the \emph{best-response mapping}: 
\begin{equation}
\label{eq:sc_best_response}
x^f = \Delta_{f}(x) \triangleq \argmax_{x'\in \mathcal{X}} f(x') - c(x, x')
\end{equation}
where $c(x, x')$ is a cost function that determines the cost of changing $x$ to $x'$,
and is assumed to be known to all.
The goal of learning is to
minimize the expected 0-1 loss, but
under the strategically-induced distribution:
\begin{equation}
\label{eq:sc_learning_objective}
\min_{f \in F} \expect{p(x, y)}{\ind{f(x^f) \neq y}}
\end{equation}
We focus on the common choice of linear classifiers
$\yhat = f(x)=\sign{(w^\top x+b)}$ and generalized quadratic costs 
$c_Q(x,x') = (x'-x)^\top Q (x'-x) = \|x'-x\|_Q^2$
for PSD $Q$.\looseness=-1

\subsection{Causal strategic classification}
A key assumption in standard strategic classification
is that changes in $x$ (via $\Delta_{f}$) do \emph{not} affect $y$;
this is encoded directly in Eq. \eqref{eq:sc_learning_objective}.
We will be interested in breaking this assumption
by allowing changes in $x$ to causally affect $y$.
To account for causal effects, we require a concrete structure that determines how changes in $x$ translate to changes in $y$.
We seek to take the minimally-necessary step for generalizing the standard setting to include causal effects.

\begin{figure}[t!]
    \centering
    \subfloat[]{\begin{tikzpicture}[node distance = 1.1cm]
            \node[circle, draw, text centered] (x_c) {$x_c$};
            \node[circle, draw, below of = x_c, minimum size=0.7cm] (U) {$u$};
            \node[circle, draw, below of = U] (x_r) {$x_r$};
            \node[circle, draw, right of = U, text centered, minimum size=0.7cm] (Y) {$y$};
    
            \draw[->, line width = 1] (x_c) -- (Y);
            \draw[->, line width = 1] (U) -- (Y);
            \draw[<->, line width = 1, dash dot] (x_r) .. controls +(left:10mm) and +(left:10mm) .. (U);
        \end{tikzpicture}}
    \hspace{1.3\baselineskip}
    \subfloat[]{\begin{tikzpicture}[node distance = 1.5cm]
            \node[circle, draw, text centered] (x_c) {$x_c$};
            \node[circle, draw, below right of = x_c, text centered, minimum size=0.7cm] (Y) {$y$};
            \node[circle, draw, below left of = Y] (x_r) {$x_r$};
    
            \draw[->, line width = 1] (x_c) -- (Y);
            \draw[<->, line width = 1, dash dot] (x_r) -- (Y);
        \end{tikzpicture}}
    \hspace{1.3\baselineskip}
    \subfloat[]{\begin{tikzpicture}[node distance = 1.5cm]
            \node[circle, draw, text centered] (x_c) {$x_c$};
            \node[circle, draw, below right of = x_c, text centered, minimum size=0.7cm] (Y) {$\yhat$};
            \node[circle, draw, below left of = Y] (x_r) {$x_r$};
    
            \draw[->, line width = 1] (x_c) -- (Y);
            \draw[->, line width = 1] (x_r) -- (Y);
        \end{tikzpicture}}
     
    \caption{
    \textbf{(a)}
    The true causal graph over user features and labels. Solid lines represent direct causal effects, dashed lines depict correlation (whose source is abstracted away).
    \textbf{(b)}
    The causal graph, as perceived by the system. Note that $u$ is unobserved, but its relation to $y$ carries over to $x_r$, making it predictively informative of $y$.
    \textbf{(c)} 
    The causal graph, as perceived by users, who seek positive predictions $\yhat=1$.\looseness=-1
    }
    \label{fig:causal_graph}
\end{figure}
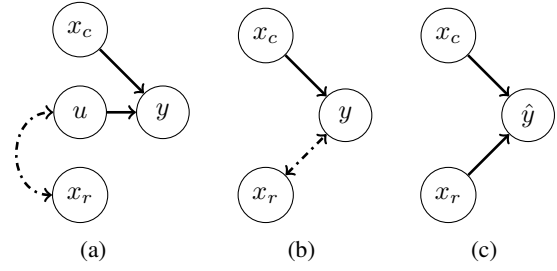

\paragraph{The causal structure.}   
Our main structural assumption is that observable features $x$ can be partitioned into
\emph{causal} features $x_c\in \mathcal{X}_c= \mathbb{R}^{d_c}$ that affect $y$, and \emph{correlative} features $x_r \in \mathcal{X}_{r}=\mathbb{R}^{d_{r}}$ that do not. \extended{\footnote{\blue{Similar assumptions have been made in 
\tocitec{all papers that assume this}; they, however, also assume that $h^*$ is linear.}}}
Together, we denote $x=(x_c,x_r)$.
To enable both $x_c$ and $x_r$ to be distinctly important in prediction,
we allow for additional \emph{unobserved} causal features, $u\in \mathcal{U}=\mathbb{R}^{d_u}$,
with which $x_r$ correlates.
Thus, $x_r$ can be informative of $y$ beyond what is conveyed by $x_c$,
and therefore 
complementarily useful in learning. 
This mimics a setting in which some known causes are observed ($x_c$), 
but alone cannot fully explain $y$,
and so are complemented by additional features ($x_r$)
which relate to other possible causes of $y$, but are themselves non-causal.
We assume $(x,u) \sim p(x,u)$ for some unknown distribution $p(x,u)$.
For labels, we consider $y$ as determined jointly by $x_c$ and $u$ via
$y \sim h^*(x_c,u)$, for some stochastic ground-truth labeling function $h^*$.
Figure~\ref{fig:causal_graph} compactly describes 
our proposed causal structure using a simple causal graph.\footnote{Despite our structural assumptions, our setting remains quite flexible.
First, relations between $x_r$ and $u$ can be arbitrary;
e.g., $x_r$ can be a causal child of $u$,
or $x_r$ and $u$ have a common parent $z$.
Second, $h^*$ can be any stochastic function of $x_c$ and $u$, and we make no assumptions on its form or relation to $F$.
Third, we allow $u$ and $x$ to be dependent (this is abstracted away in Fig.~\ref{fig:causal_graph}).
Fourth, we assume $u$ includes \emph{some} variables that correlate with $x_r$, but make no assumptions on, nor require knowledge of, their nature.\looseness=-1
} 

\extended{
\vspace{0.5em}
\begin{remark}
Our approach in Sec. \ref{sec:method}
relies operationally on defining a  
partitioning of observed features into $x_c$ and $x_r$.
This mimics a setting in which there are some observable known causes ($x_c$), 
but which alone cannot fully explain $y$, and so are complemented by additional features ($x_r$) which relate to other possible causes of $y$,
but are themselves non-causal.
\blue{In Sec. \toref\ we elaborate on why and how such knowledge can be useful,
and point at the risks that surface when it is unaccounted for
(e.g., by {\naive}ly considering features to be either all correlative, or all causal)}.\looseness=-1
In principle, determining the feature partition can be done either using expert knowledge, through causal discovery \citep{glymour2019review},
or when conditions permit, via sequential learning  \citep{bechavod2021gaming}. \extended{\footnote{Note that inferring the partition requires only partial information regarding the graph.}}
We view this effort as orthogonal to our goals,
and here focus primarily on optimizing accuracy under a given partition,
but also empirically explore the implications of learning under imperfect knowledge (Sec. \toref).
\todo{think carefully if to include the second half about inference - risky!!}
\end{remark}
}

\begin{figure}[t!]
    \centering
    \includegraphics[width=0.95\columnwidth,trim={0.6cm 0 0 0},clip]{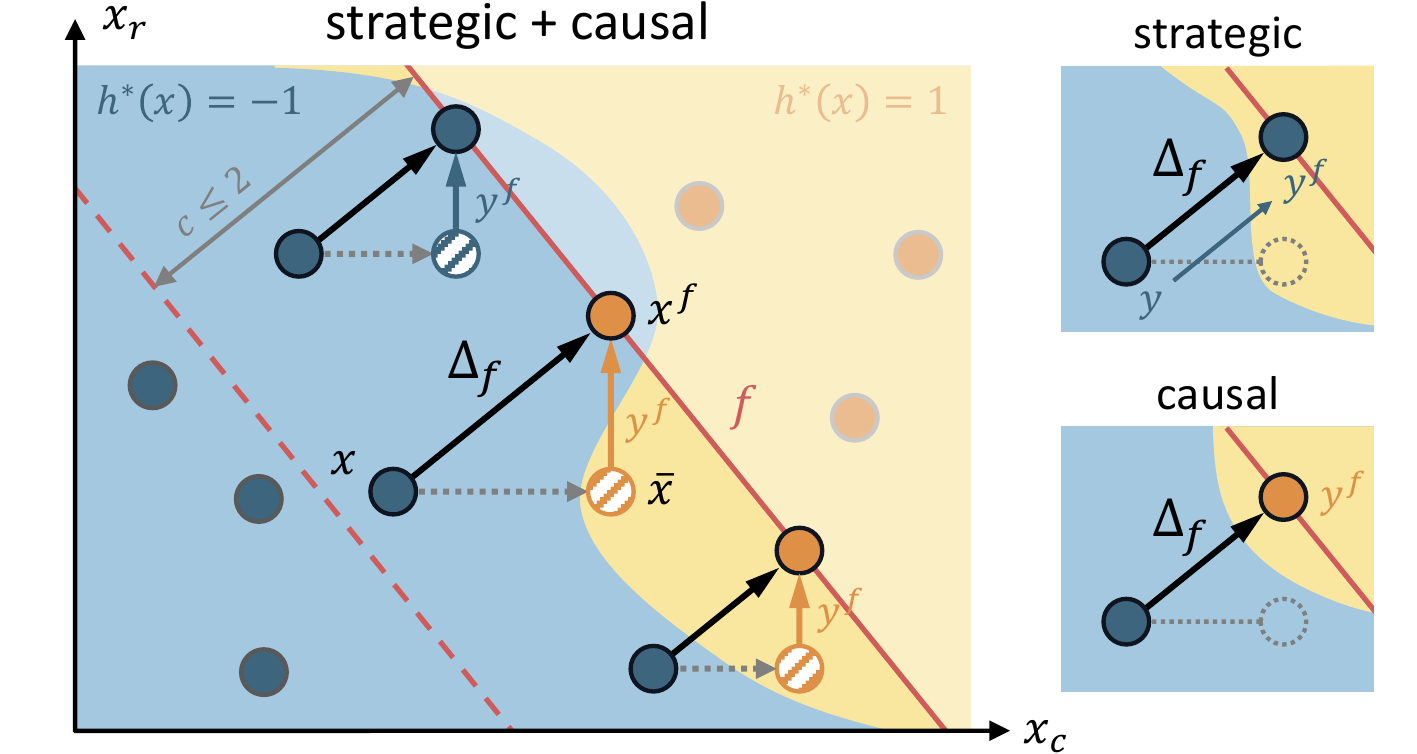}
    \caption{
    \textbf{An example of causal strategic classification.}
    Given $f$, strategic users move clean points $x=(x_c,x_r)$ onto the decision boundary,
    $x^f = \Delta_f(x)$,
    when costs permit.
    Labels, however, are affected only by the updated causal component, $x^f_c$, via $y^f=h^*(x^f_c,u)$
    (c.f. strategic-only or causal-only cases).
    For $x_r=u$, labels $y^f$
    are given by projecting $x^f$ onto the causal subspace, $\xbar=(x^f_c,x_r)$.\looseness=-1
    }
    \label{fig:projection_illustration}
\end{figure}

\paragraph{Implications on learning.}
Since only $(x_c,x_r)$ are observed, the classifier $f$ can only be a function of these, i.e., $f(x_c,x_r)$,
and users respond just as in
Eq. \eqref{eq:sc_best_response}, i.e., via:
\begin{equation}
x^f = (x^f_c,x^f_r)=\Delta_f(x_c,x_r)
\end{equation}
Note this implies that users are incentivized to change  only
$x_c$ and $x_r$, but \emph{not} $u$:
\extended{Given this fact, together with the causal reliance of $y$ on $x_c$ (and $u$), we get that}
once $\Delta_f$ has been applied,
the underlying features become $(x^f_c,x^f_r, u)$,
and the updated label---which is the true target of prediction---is $y^f = h^*(x^f_c,u)$.
Given this, our causal strategic learning objective is:\looseness=-1
\begin{equation} \label{eq:causal_sc_learning_objective}
    \min_{f \in F} \expect{p(x,u)}{\ind{f(x^f) \neq h^*(x_c^f, u)}} 
\end{equation}
In the simple case where $x_r=u$
(but noting $x_r{\mapsto}x^f_r$ does \emph{not} change $u$),
$y^f$ can be interpreted as $h^*(\xbar)$,
where $\xbar = (x^f_c,x_r)$ is the `projection' of $x^f$
onto the causal subspace 
(see Figure~\ref{fig:projection_illustration}).
In the special case where there are no causal features (i.e., $x=x_r$),
Eq.~\eqref{eq:causal_sc_learning_objective}
reduces to the standard
strategic classification objective in Eq.~\eqref{eq:sc_learning_objective}
with $y=h^*(u)$, and any discrepancies between $x_r$ and $u$ manifest as noise.

\paragraph{Challenges and prospects.}
The main challenge in optimizing Eq.~\eqref{eq:causal_sc_learning_objective}
is that in addition to accounting for strategic responses $x \mapsto x^f$,
learning must now also anticipate how such changes affect labels
via $y^f=h^*(x^f_c,u)$.
Intuitively, since points move to obtain positive predictions $\yhat=1$, 
correctly estimating $y^f$ is important for
(i) \emph{avoiding} negative post-strategic labels $y^f=-1$ (on which $f$ errs),
as well as for (ii) \emph{encouraging} positive post-strategic labels $y^f=1$ (on which $f$ is correct).
This reveals how user interests ($\yhat=1$) align system goals ($\yhat=y^f$) with the general aim of improvement ($y^f=1$).
Nonetheless, these notions remain distinct, 
and optimizing for one criterion 
does \emph{not} imply optimality for the other 
(see Appendix \ref{apx:acc_vs_improve}).

Since $h^*$ is unknown,
our
approach for optimizing 
Eq.~\eqref{eq:causal_sc_learning_objective}
will be to replace $h^*$ with some
estimated $\hhat$.
However, since training data $S$ includes only `clean' points $(x,y)$,
the challenge in this is twofold:
(i) due to strategic behavior, $h^*$ might need to be queried on points that lie outside the data distribution, and for which $S$ (on which $\hhat$ is trained) may not be representative,
and (ii) even though $x^f_c$ can be computed, $u$ remains to be unobserved.
As we will show, allowing learning to make use of additional
`dirty' data $(x^f,y^f)$,
collected \emph{over time} and under different deployed classifiers $f$,
can enable learning to contend with these challenges.\looseness=-1




\subsection{Learning over time}\label{sec:learning_over_time}
To study
temporal aspects of causal strategic learning,
we adopt the general formulation of \emph{performative prediction} \citep{perdomo2020performative}.
Here, learning proceeds in discrete rounds,
where at each round $t>0$
the currently deployed model $f_t$ determines the data distribution
in the next round, $p_{t+1}=p^{f_t}=D(f_t;p_0)$, for some initial $p_0$ and distribution mapping $D$.
The overall goal is 
to optimize $f$ on the distribution it induces,
namely minimize the \emph{performative risk}:\looseness=-1
\begin{equation}
\label{eq:perf_risk}
\min_{f \in F} \expect{(x,y) \sim p^f}{\ind{f(x) \neq y}}
\end{equation}
In our setting, $D$ corresponds to feature updates via $\Delta_f$
and label updates via $h^*$,
and Eq. \eqref{eq:causal_sc_learning_objective} 
is a special case of Eq. \eqref{eq:perf_risk}.\footnote{Note Eq.~\eqref{eq:sc_learning_objective} is also a special case of Eq.~\eqref{eq:perf_risk},
but which is known to converge after one round when $\Delta$ is known.}\looseness=-1

Similarly to \citet{miller2021outside}, we allow $T$ rounds of retraining and deployment.
At each round $t \le T$, 
the system observes new data $S_t \sim p_t$,
and (re)-trains $f_t$;
then, it deploys $f_t$, which induces a distribution shift: 
\[
S_t \sim p_t \, \overset{train}{\longrightarrow} \, f_t \, \overset{induce}{\longrightarrow} \, p^{f_t}=p_{t+1}, \qquad
p_0=p
\]
where in our setting $p_0 = p$ is the `clean' distribution,
and $S=S_0$.
In retraining dynamics, each $f_t$ is optimized for accuracy using currently available data:
this relates to settings where deployed models are used throughout the dynamics, and so are required to perform well at each point in time.
Note rounds $t>0$ includes fresh samples,
which consist purely of `dirty' inputs $(x^{f_t},y^{f_t}) \sim p_{t+1}$;
importantly, for these points, their corresponding original `clean' $(x,y)$ remain unknown.
Finally, at time $T$, the system commits to some final $f \in \{f_t\}_{t < T}$, 
to be used henceforth,
and on which performative risk is evaluated.\looseness=-1

\extended{
The general premise in performative prediction is that
at each round, $f_t$ should be trained to obtain high predictive accuracy,
which relates to settings where deployed models are used throughout the dynamics and so are required to perform well at each point in time.
This justifies the notion of retraining, and restricts arbitrary exploration.
Here we follow suit, and train each $f_t$ to predict well on the induced $p^{f_t}$.
As time proceeds, we can expect additionally gathered data to be helpful in reducing uncertainty about $h^*$, and consequently,
in performative risk. 
We next present analysis which sheds light on when and how this can occur.
}


\section{Analysis} \label{sec:analysis}



To learn well in causal strategic settings, we must first
understand how strategic behavior and causal effects translate into distribution shifts.
In this section we characterize such shifts
by analyzing the induced marginal $p^f(x^f)$ and conditional $p^f(y^f|x)$ for different cases.
%
%
Throughout we use capital letters to denote random variables (e.g., $X,U,Y$) and lowercase for their realizations (e.g., $x,u,y$).
Our analysis makes use of an `inverse' response mapping operator:
\begin{equation}
\Delta_{f}^{\minus 1}(x') \triangleq \{x \mid \Delta_{f}(x)=x'\}
\end{equation}
which for any `shifted' point $x'$ returns the set of points $x$ from which $x'$ could have originated. 
For simplicity here we present results for deterministic $h^*$,
but these also hold in the stochastic case.
Proofs are deferred to Appendix~\ref{apx:proofs}.
\looseness=-1



\subsection{Case \#1: Correlative-only features}
Using $x_{r}$ to predict $y$ can be useful due to its correlation with $u$, which is a direct (and potentially distinct) cause of $y$.
When $f$ relies only on $x_{r}$, all of the features that are used for learning are non-causal, hence changes in $x$ do not affect $y$.\looseness=-1
\begin{observation} \label{obs:using only x_r}
When using only $x_{r}$, causal strategic classification reduces to standard strategic classification.
\end{observation}
Denote $x_{r}'=\Delta_{f}(x_{r})$.
Our first result shows the connection between
the original and induced distributions.\looseness=-1
\begin{lemma}
\label{lem:only_xr}
Let $p(x_r,y)$ be some base distribution.
Then for any classifier $f$, the induced $p^f(x'_r,y)$
can be expressed using the base
marginal $p(x_{r})$ and conditional $p(y|x_{r})$ as:\looseness=-1
\begin{align}
p^f(x_{r}') &=
\int_{x_{r} \in \Delta_{f}^{\minus 1}(x_{r}')}p(x_{r}) \d x_r,
\label{eq:only_xr_marginal} \\
p^f(y|x_{r}') &=
\int_{x_{r} \in \Delta_{f}^{\minus 1}(x_{r}')}
\frac{p(x_{r})}{p^f(x_{r}')} p(y|x_{r}) \d x_r
\label{eq:only_xr_conditional}
\end{align}
\end{lemma}
Eq. \eqref{eq:only_xr_marginal} simply states that the probability of observing some modified $x'_r$ derives from all points $x_r$ that map to it via $\Delta_f$.
Eq. \eqref{eq:only_xr_conditional} then shows that predicting for $x'_r$ its corresponding $y$ (which remains unmodified) requires reasoning about the possible labels of all points in $\Delta^{\minus 1}_f(x'_r)$;
since this is a set, the implication is inherent (informational) uncertainty in $y$,
which cannot be reduced through statistical means (i.e., observing more data from $p^f$).
This reveals the mechanism through which strategic behavior can hinder accuracy,
where
$\frac{p(x_{r})}{p^f(x_{r}')}$ expresses
how strategic behavior `distorts' the base probability $p(y|x_{r})$
(which already includes any uncertainty due to $u$, and to $x_c$ if it exists).

Lemma \ref{lem:only_xr} shows that
the case of $x=x_r$ entails
\emph{full distribution shift}, i.e., 
both $p(x)$ and $p(y|x)$ can vary.
Nonetheless, it provides useful insight,
which is immediate from Eq.~\eqref{eq:only_xr_conditional}:\looseness=-1
\begin{corollary} \label{cor:using only x_r}
When using only $x_{r}$,
knowing the base distribution suffices for constructing
the Bayes-optimal classifier.
\end{corollary}
Corollary \ref{cor:using only x_r} highlights why
clean samples are useful;
it also suggests that learning might benefit from incorporating a marginal density estimator, $\phat(x)$, into the training procedure---a notion we adopt in our algorithm in Sec.~\ref{sec:method}.\looseness=-1

\subsection{Case \#2: Causal-only features}
Using $x_{c}$ is useful for learning as it is a direct cause of $y$ in itself.
We now analyze the case of using only $x_c$ for prediction,
which requires us to directly account for $u$.
In this case, the base conditional has the following form:
\begin{align}
\label{eq:base_conditional}
p(y|x_c) 
&= \int_{u}p(u|x_c)p(y|x_c, u) \d u \nonumber \\
&= \int_{u}p(u)\cdot \mathbbm{1}{\{y= h^*(x_c, u) \}} \d u \nonumber \\
&
= \int_{u:\:y= h^*(x_c, u)}p(u) \d u
\end{align}
which is simply the uncertainty in $y$ due to $u$.
Further assuming that $X_c$ and $U$ are independent
reveals a tight connection.
Denote $x_{c}'=\Delta_{f}(x_{c})$ and $y'=h^*(x_{c}', u)$.
\begin{lemma}
\label{lem:only_xc}
Let $p(x_c,y)$ be some base distribution,
and assume $x_c{\perp}u$.
Then for any classifier $f$, we have:
\begin{align}
p^f(y'|x_c') 
= \int_{u:\:y'= h^*(x_c', u)}p(u) \d u
\label{eq:only_xc_conditional}
\end{align}
\extended{add steps}
and $p^f(x_{c}')$ is as in Eq.~\eqref{eq:only_xr_marginal}
(with $x_c$ replacing $x_r$).
\end{lemma}
Because the induced marginal is susceptible only to strategic effects,
its form remains the same regardless of which features are used.
More interestingly, 
Eq.~\eqref{eq:only_xc_conditional} states that 
the induced conditional $p^f(y'|x'_c)$
remains exactly the same as the \emph{original} base $p(y|x)$ (Eq.~\eqref{eq:base_conditional}).
Thus, causal effects `cancel out' the strategic effect of $f$ on $p(y|x_c)$,
and only $p^f(x_c)$ remains susceptible to strategic shifts.
Thus,
for \emph{any} $f$,
it holds that $P(Y'=y'|X_{c}'=x_{c}')=P(Y=y'|X_{c}=x_{c}')$,
which is a 
special case of \emph{covariate shift} \cite{shimodaira2000improving}.
\begin{corollary} \label{cor:using only x_c when x,u independent}
If $x_c$ and $u$ are independent, then irrespective of $f$, using only $x_{c}$ reduces to learning under covariate shift.
\end{corollary}

\subsection{Case \#3: Using all features}
We now consider the most general case when all feature types are used
(and with no assumptions on independence).
\begin{lemma}
For any $p(x,u)$ and any classifier $f$, we have:
\label{lem:xc_and_xr}
\begin{align}
    & p^f(y'|x') = 
    \int_{u:\:y'= h^*(x'_{c}, u)}
    \nu_f(u;x') p(u) \d u \label{eq:only_xc_dep_conditional} \\
    \intertext{where:}
    & \nu_f(u;x') = 
    \int_{x \in \Delta_{f}^{\minus 1}(x')}
    \frac{p(x | u)}{p^f(x')} \d x \label{eq:only_xc_dep_nu}
\end{align}
and $p^f(x')$ is as in Eq.~\eqref{eq:only_xr_marginal}
(with $x$ replacing $x_r$).
\end{lemma}
While covariate shift no longer holds,
note that Eq.~\eqref{eq:only_xc_dep_conditional} matches
Eq.~\eqref{eq:only_xc_conditional} up to the term $\nu_f$.
Hence, $\nu_f$
quantifies the deviation
from covariate shift due to $f$:
when $\nu_f(u;x')$ takes values close to one across $u$,
then covariate shift `approximately' holds;
otherwise, we have a particular form of full distribution shift.
Note that in itself, $\nu_f$ relates to Eq.~\eqref{eq:only_xr_conditional},
in that the strategic effects also express as a distorted probability term integrated over the inverse response set.\looseness=-1


\extended{\todo{think if can connect this to bound on generalization error}}


\paragraph{Interpretation.}
Our analysis thus far reveals a tradeoff:
Correlative features $x_r$ are susceptible to gaming---which
manifests as full distribution shift,
but requires only clean data to accommodate.
Conversely, causal features $x_c$ (and their relation to $u$) bring learning closer to covariate shift, 
which is simpler,
but introduces larger uncertainty in $y$.
Note this uncertainty stems from points $x^f$ moving to
regions of low density under $p$;
hence, in principle, dirty data gathered over time and in response to different models $f$ may aid in decreasing uncertainty and improving performance.
Our approach,
presented next,
aims to balance these two forces.

\section{Method} \label{sec:method}


Recall that our goal is to optimize the causal strategic learning objective in Eq. \eqref{eq:causal_sc_learning_objective}.
Given a finite sample $S$, 
we adopt the conventional ERM approach
and aim to minimize the empirical risk. 
Ideally, we would like to solve:
\begin{equation} \label{eq:erm_objective}
    \argmin_{f \in F} \sum\nolimits_{(x,y) \in S}
    \ind{f(x^f) \neq h^*(x_c^f,u)}
\end{equation}
However, this introduces several challenges:
(i) the 0-1 loss is non-differentiable;
(ii) $x^f$ is the output of $\Delta_f(x)$, which is an argmax operator
that is also non-differentiable;
(iii) $h^*$ is unknown,
and (iv) $u$ is unobserved,
which together prevent us from computing updated labels $y^f=h^*(x_c^f,u)$.
Also note that Eq. \eqref{eq:erm_objective} makes no use of the observed clean $y$.\looseness=-1

Our first step is to replace $\ind{\cdot}$ with an appropriate proxy loss;
for this, we adopt the \emph{strategic hinge} $\shinge$ from \citet{levanon2022generalized},
which accounts for strategic behavior
and provides favorable generalization guarantees.
Importantly, it does not explicitly rely on $\Delta_f$,
and is entirely differentiable.
Next,
for handling $h^*$ and $u$,
our general approach will be to replace $h^*$ with a learned $h$, and use $x_r$ as a surrogate for $u$.\footnote{Our approach requires to operationally define a feature partition as input to the learning algorithm.
In Sec.~\ref{sec:exp_sensitivity} we empirically demonstrate
its robustness to misspecified partitionings.}
We first describe our approach for clean data,
and then extend it to utilize additional dirty data.
Pseudocode for our entire procedure
is given in Algorithm (\ref{alg:cserm}).\looseness=-1

\subsection{Learning with clean data}
We propose to replace $h^*$ in Eq. \eqref{eq:erm_objective}
with a differentiable estimate 
$h: \mathcal{X}_c \times \mathcal{X}_{r} \rightarrow [-1,1]$,
learned from data over some chosen function class $H$.
Here the goal is to exploit the correlation between $u$ and the observed (pre-strategic) $x_{r}$;
i.e. $h$ uses $x_{r}$ as a "substitute" for how $h^*$ uses $u$,
and as a complement to $x_c^f$.\footnote{Note $h^*$  
takes inputs $(x_c,u)$, whereas $h$ operates on $(x_c,x_r)$. \looseness=-1
\extended{
a function of $x_c$ and $u$, whereas $h$ is a function of $x_c$ and $x_r$. Generally, $u$ and $x_r$ can be entirely different objects.
}
}
Since clean data includes clean labels $y=h^*(x_c,u)$,
we can use $S$ to optimize $h$ via: 
\begin{equation}
h = \argmin_{h' \in H} \sum\nolimits_{(x,y) \in S}
\loss \left( h'(x_c, x_r), y \right)
\end{equation}
for some standard proxy loss $\loss$ (e.g., hinge loss or log loss).
Note that $h$ is learned on $(x_c,x_r)$, but used on $(x_c^f,x_r)$.
In principle, $h^*$ is needed only for points that move,
since if $x^f \neq x$ then $y^f=h^*(x_{c}^f, u)$,
and otherwise $y^f = y$.
To prevent $h$ from
needlessly erring in such cases, we implement $y^f$
as a  differentiable `soft if', $\ytilde_h$, as follows.
First, note that $x$ moves iff it is necessary \emph{and} cost-effective, i.e.:\looseness=-1
\begin{equation*}
\ind{x \neq x^f} = \ind{f(x)=-1} \cdot \ind{c(x,x^f)<2}
\end{equation*}
Next, we relax this and define a soft movement indicator:
\begin{equation*}
\mu(x, x^f) = 
2 \left( \sigma_{\tau}(c(x, x')) -1/2\right) \cdot
\sigma_{\tau}\left(2-c(x, x^f)\right)
\end{equation*}
where $\sigma$ is a sigmoid with temperature $\tau$.
Then, we express $y^f$ using $h$ and $\mu$,
which gives our soft updated label:
\begin{equation}
\ytilde_h(x,x^f) = 
    y + \left(h(x_c^f, x_r) - y\right)\cdot \mu (x, x^f)
\end{equation}
Finally, given $h$, our proposed objective for clean data is:
\begin{equation}
\label{eq:clean_objective}
\argmin_{f \in F} \sum\nolimits_{(x,y) \in S}
\shinge \left( f(x^f), \ytilde_h(x,x^f) \right)
\end{equation}


\begin{algorithm}[tb]
   \caption{CSERM}
   \label{alg:cserm}
\begin{algorithmic}[1]
    \STATE {\bfseries Input:} clean data $S$, regularization schedule $\lambda_t$
    \STATE $S_0\gets S$
    \STATE $\phat\gets \text{KDE}(S)$
    \FOR{$t=0,\ldots T-1$}
        \STATE update $S_{\le t}$ to include $S_t$
        \STATE $\qhat\gets \text{KDE}(S_{\le t})$
        \STATE $h_t \gets \argmin_{h\in \mathcal{H}} \sum_{(x,y)\in S_{\le t}}{\loss\left(h(x), y\right) }$
        \STATE $f_t \gets \argmin_{f \in F} \sum_{(x,y) \in S}\shinge \left( f(x^f), \ytilde_{h_t}(x,x^f) \right)$
        \myindent{4} $+ \lambda_t R(f;S,\qhat)$
        \STATE publish $f_t$ and collect dirty samples $S_{t+1}\sim p^{f_{t}}$
        \FOR{$(x_c^{f_t},x_r^{f_t},y^{f_t})\in S_{t+1}$}
            \STATE $\xtilde_r \gets \expect{x_r \sim \phat(x_r | x^{f_t})}{x_r}$
            \STATE replace $x_r^{f_t}$ with $\xtilde_r$
        \ENDFOR
    \ENDFOR
\end{algorithmic}
\end{algorithm}


\subsection{Utilizing additional dirty data}
The limitation in using only clean data for training $h$
is that $h$ is tailored to $p$, and may not approximate $h^*$ well outside it---a likely scenario when points move strategically.
Towards this,
we propose to use temporally-gathered dirty data,
sampled from induced distributions $p^f$,
to iteratively improve $h$ by retraining it at each round $t$ on all available data,
namely training $h_t$ on $S_{\le t}=\cup_{t' \le t} S_{t'}$ where $S_{t'} \sim p^{t'}=p^{f_{t'-1}}$. 
Unfortunately, dirty data isn't immediately useful,
and {\naive}ly training $h_t$ on it may introduce bias.
To see why, note that dirty data includes inputs $(x^f_c,x^f_r,y^f)$;
in contrast, the observed $y^f$ depends via $h^*$ on $x^f_c$ and on $u$.
Whereas $x^f_c$ is useful, what we require is the original $x_r$, which is informative of $u$---instead, we observe the modified $x^f_r$, which is inappropriate.
Ideally, we would like to train $h_t$ on `mixed' pairs $(x^f_c,x_r)$,
but these are unavailable.
As a solution, 
we propose to reconstruct $\xtilde_r \approx x_r$ using a density model $\phat(x) \approx p(x)$,
trained once at the onset on clean data.
Since $x^f$ is the strategic response to some $x$,
we can estimate the likelihood for any given $x_r$
as $\phat(x_r|x^f)$;
by considering all points in $\Delta^{\minus 1}_f(x^f)$,
we define:\looseness=-1
\begin{equation*}
    \phat(x_r|x^f) \triangleq \frac{1}{\phat^f(x^f)}\int_{x_c: (x_c,x_r)\in \Delta_f^{\minus 1}(x^f)} \phat(x_c,x_r)\d x_c
\end{equation*}
To obtain a single entry, we compute the expected value:
\begin{equation}
\label{eq:xr_tilde}
\xtilde = (x^f_c,\xtilde_r), \,\,\text{where}\,\,
\xtilde_r = \expect{x_r \sim \phat(x_r | x^f)}{x_r}
\end{equation}
and use these for training $h$.
Appendix~\ref{apx:efficient_computation} shows how 
to efficiently compute Eq.~\eqref{eq:xr_tilde} ,
using the fact that $\Delta^{\minus 1}_f(x^f)$
can be expressed as a closed interval of points in $\R^d$.\looseness=-1

\paragraph{Regularization for exploration.}
Although dirty data can be helpful in extending the regions of data on which $h_t$ is trained,
\emph{what} those regions are is determined entirely
by the set of previous $f_t$.
To promote variation in dirty data,
we propose to augment Eq.~\eqref{eq:clean_objective} with a regularization term
that encourages $f$ to push points $x^f$ to regions of low density:
\begin{equation}
\label{eq:regularizer}
R(f;S,\qhat) = \frac{1}{|S|} 
\sum\nolimits_{x \in S} \log \, \qhat(x^f_c, x_r)
\end{equation}
Here, $\qhat$ is a density model that is
(re)-trained on aggregate data $S_{\le t}$ at each round $t$,
since its role is to inform us of uncertainty in $h_t$.
Our final regularized learning objective is:\looseness=-1
\begin{equation}
\label{eq:regularized_objective}
\argmin_{f \in F} \sum_{(x,y) \in S}
\shinge \left( f(x^f), \ytilde_h(x,x^f) \right)
\,+\, \lambda R(f;S,\qhat)
\end{equation}
This equips our approach with a mild form of exploration,
whose degree is determined by $\lambda$.
Practically we found it useful to use a gradually decaying $\lambda_t$:
this places initial emphasis on exploration,
which gradually shifts towards exploitation as more data is collected.
In our experiments we use a kernel density estimator (KDE)
for $\qhat$, which is differentiable;
hence, the entire objective can be trained end-to-end.



\section{Experiments using synthetic data}
We begin with a series of synthetic experiments,
each designed to demonstrate a different aspect of our setting and approach.
We consider $x \in \R^2$ (which can be visualized),
and fix $x_1=x_c$ and $x_2=x_r=u$.
We compare learning using our strategic causal approach (\CSERM) to 
(i) a \naive\ \ERM\ approach,
and (ii) a strategically-aware (but causally-oblivious) baseline that optimizes Eq. \eqref{eq:sc_learning_objective} using the approach in \citet{levanon2022generalized} (\SERM).
We also consider a non-strategic benchmark (\bench) in which \ERM\ is evaluated on clean (i.e., non-strategic) data.\looseness=-1

\begin{figure}[t!]
    \centering
    \includegraphics[width=\columnwidth]{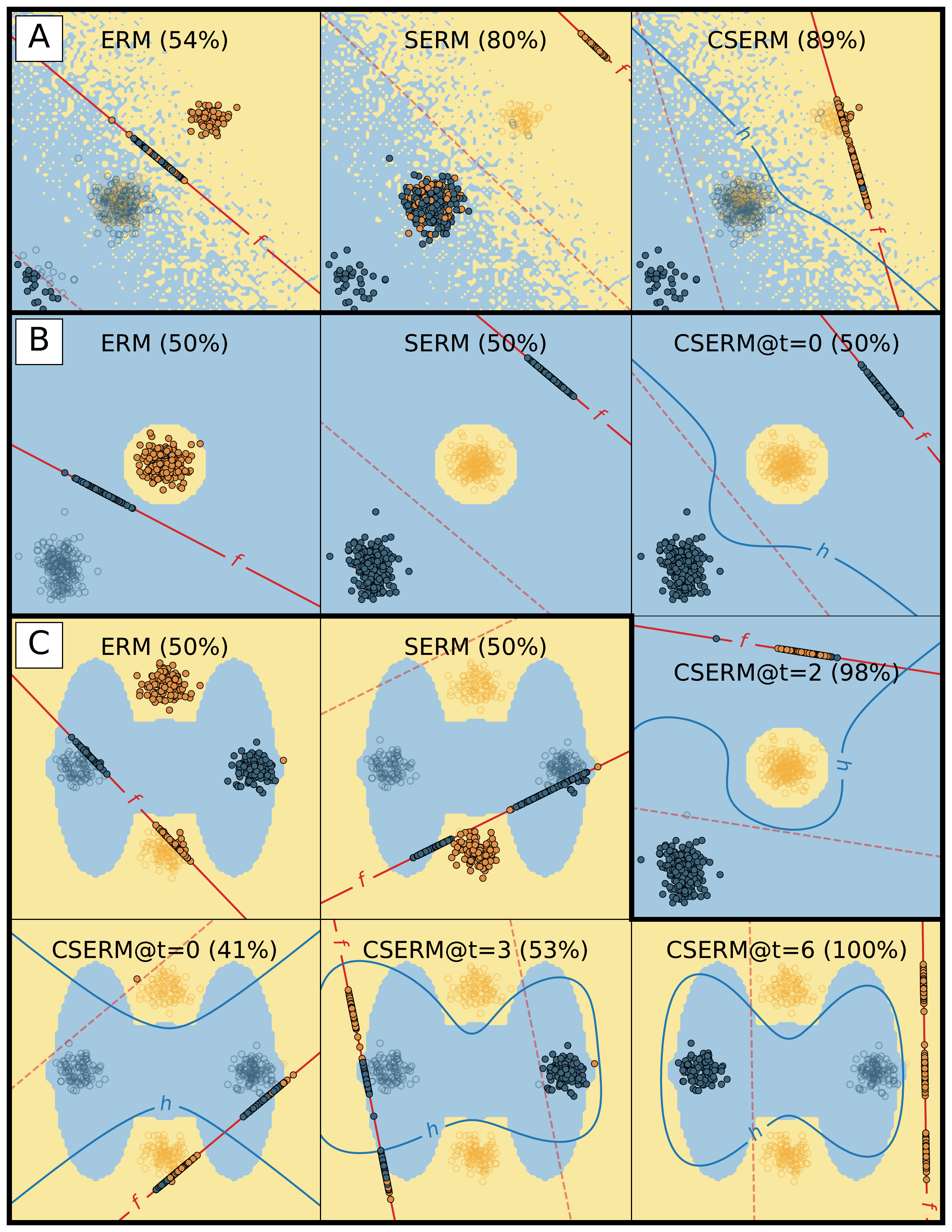}
    \caption{
    Results on synthetic experiments (A,B,C) for \ERM,
    strategic \ERM\ (\SERM),
    and our approach (\CSERM)
    (acc. in parentheses).
    Here $x_1{=}x_c$ (x-axis) and $x_2{=}x_r{=}u$ (y-axis).
    Hollow circles show pre-modified (`clean') data,
    filled circles show strategically-modified (`dirty') data.
    Colored regions depict $h^*$;
    red lines show learned $f$, dashed lines mark
    regions of movement for $\Delta_f$,
    blue lines show learned $h$.\looseness=-1
    }
    \label{fig:syth}
\end{figure}

\paragraph{Utilizing improvement.}
Our first experiment studies the ability of our approach
to identify and make use of regions where $h^*$ is \emph{positive}
to increase predictive performance (Fig.~\ref{fig:syth} A).
We construct $p(x_c,u)$ to include two clusters that are separable by a linear $h^*$,
but inject noise so that it is no longer separable by any $f(x)$
(while preserving the majority class in each cluster).
As expected, \SERM\ (80\% accuracy)
operates by taking the optimal \ERM\ solution (54\%)
and making it more strict to prevent negative points from crossing;
from its own perspective, this is sensible,
since if $y$ does not change, then negative points that move cause $f$ to err.
In contrast, \CSERM\ 
(89\%)
utilizes its knowledge of $h^*$
(via $h$)
to push negative points to positive regions; once these points move,
they obtain both positive predictions \emph{and} positive labels,
and accuracy increases---surpassing \bench\ (80\%).\looseness=-1

\begin{table*}[t!]
\centering
\caption{Results on real data.}
\resizebox{\textwidth}{!}{
\begin{tabular}{lrccccccccccc}
  &   & \multicolumn{5}{c}{\textbf{\texttt{card fraud}}} &   & \multicolumn{5}{c}{\textbf{\texttt{spam}}} \\
\cmidrule{3-7}\cmidrule{9-13}  &   & accuracy & \%improve & \%move & \%neg${\mapsto}$pos & welfare &   & accuracy & perceived & \%improve & \%move & \%pos${\mapsto}$neg \\
\cmidrule{1-1}\cmidrule{3-7}\cmidrule{9-13}\CSERMlambda &   & \boldmath{}\textbf{87.8\tiny{\,$\pm0.2$}}\unboldmath{} & \textbf{12.2} & 60.1 & \textbf{13.8} & -0.65 &   & \boldmath{}\textbf{92.7\tiny{\,$\pm0.5$}}\unboldmath{} & 97.0 & 3.1 & 37.5 & \textbf{0.1} \\
\CSERM &   & \boldmath{}\textbf{86.6\tiny{\,$\pm0.5$}}\unboldmath{} & \textbf{10.2} & 58.9 & \textbf{11.8} & -0.48 &   & \boldmath{}\textbf{92.4\tiny{\,$\pm0.4$}}\unboldmath{} & 97.1 & 2.4 & 36.7 & \textbf{0.1} \\
\SERM &   & 78.4\tiny{\,$\pm0.2$} & 0.8 & 45.9 & 1.6 & -0.16 &   & 84.0\tiny{\,$\pm0.3$} & \textbf{91.2} & \textbf{-7.2} & 41.3 & 7.2 \\
\RRM &   & 75.8\tiny{\,$\pm0.5$} & 0.5 & 24.7 & 0.7 & -0.06 &   & 77.2\tiny{\,$\pm2.4$} & 76.6 & -2.1 & 30.5 & 2.8 \\
\ERM &   & 66.7\tiny{\,$\pm0.6$} & 0.3 & 19.8 & 0.4 & 0.25 &   & 75.4\tiny{\,$\pm0.3$} & 91.2 & 0.4 & 17.1 & 0.0 \\
\cmidrule{1-1}\cmidrule{3-7}\cmidrule{9-13}\CSERMoracle &   & 87.0\tiny{\,$\pm0.2$} & 10.1 & 57.9 & 11.8 & -0.60 &   & 93.5\tiny{\,$\pm0.1$} & 93.5 & 4.5 & 41.7 & 0.0 \\
\end{tabular}%
}%
  \label{tbl:real_experiments}%
\end{table*}%

\paragraph{Avoiding pitfalls.}
We next experiment in a setting in which knowledge about where $h^*$ is \emph{negative} is crucial for preserving accuracy (Fig.~\ref{fig:syth} B).
Here the clean $p(x_c,u)$ also defines two clusters,
but which can now be separated by a learned $f$.
However, outside $p(x_c,u)$,
we define $h^*$ to be positive precisely on the positive cluster, and negative elsewhere.
Here the optimal solution is to use only $x_r$ since it preserves the original $y$.
\ERM\ (50\%) fails due to strategic behavior;
\SERM\ (50\%) anticipates strategic responses,
but is oblivious to $h^*$,
and so inadvertently pushes positive points to become negative, and errs.
\CSERM, by estimating $h^*$ with $h$, is able to find the optimal solution,
though this takes time. 



\paragraph{The role of exploration.}
Our last synthetic experiment considers the canonical XOR classification task,
which is well-known to be non-linearly separable (Fig.~\ref{fig:syth} C).
Indeed, \ERM\ fails catastrophically (50\%),
as does \SERM\ (50\%).
Nonetheless, 
our approach can obtain perfect accuracy---by
utilizing causal knowledge to incentivize strategic behavior that \emph{makes} the data separable.
This, however, requires exploration:
without regularization, \CSERM\ is unable to improve,
since newly collected dirty points do not improve $h$;
however, by encouraging $f$ to uncover uncertain regions,
$h$ improves over time, until it is sufficiently informative of $h^*$
for learning to find the optimal $f$ (100\%),
which pushes one cluster of negative points to a positive region.


\section{Experiments using real data} \label{sec:exp_real}
\begin{spacing}{0.99}
We now turn to experiments based on real data 
using two public datasets:
(i) \texttt{spam}, used originally in \citet{hardt2016strategic},
and (ii) \texttt{card fraud},
used in \citet{levanon2021strategic}.
Appendix~\ref{apx:experiments_real}
includes further details on data, methods, and optimization.
\looseness=-1

\paragraph{Procedure.}
Experimenting in a causal setting requires us to be able to query
labels for arbitrary (modified) points $(x'_c,u)$.
Towards this, we begin each experiment by 
determining a partition of the original features into $x_c$ and $u$,
and use points $(x_c,u)$ to train a ground-truth labeling function $h^*$ using original labels $y^*$.
For consistency we use $h^*$ to generate labels $y$ for both clean and dirty examples.
We then define a mapping $u{\mapsto}x_r$,
which can be lossy and noisy.


Next, we split the data roughly 60-10-30
into train, validation, and test sets.
The train set is then further partitioned into a clean set,
and an inventory from which dirty data is sampled
(see Appendix~\ref{apx:vaying_cost} for additional results on different ratios of clean vs. dirty data).
Validation data is used for 
early stopping and model selection, 
and held-out test data is used for final evaluation.
In line with our temporal setup in Sec.~\ref{sec:learning_over_time},
we consider $T=10$ rounds of retraining,
where at each round $t$, 
we generate on the basis of $f_t$ dirty samples $S_t$.
These are obtained by taking a $1/T$-portion of the reserved inventory,
and simulating strategic responses via $x^{f_t} = \Delta_{f_t}(x)$
and label updates $y^{f_t} = h^*(x^{f_t}_c, u)$.
Once $S_t$ is obtained, at round $t+1$ it can be used for training $f_{t+1}$.
Non-temporal methods are given access to the full (clean) train set.
Costs are $c_\alpha(x,x')=\alpha\|x-x'\|_2^2$,
where for consistency across datasets 
we set $\alpha$
so that $\sim 50\%$ of points move on round $t=1$.
Appendix~\ref{apx:vaying_cost} includes results on additional $\alpha$-s
and for all methods,
exhibiting qualitatively similar performance trends.\looseness=-1


Finally, we run all methods and compare performance.
For methods that make use of dirty data over time,
we report results for each round,
as well as for the best model (chosen on validation data) to which the method commits.
We report average results and standard errors 
over 15 random splits.\looseness=-1


\begin{figure*}[t!]
    \centering
    \includegraphics[width=0.64\textwidth]{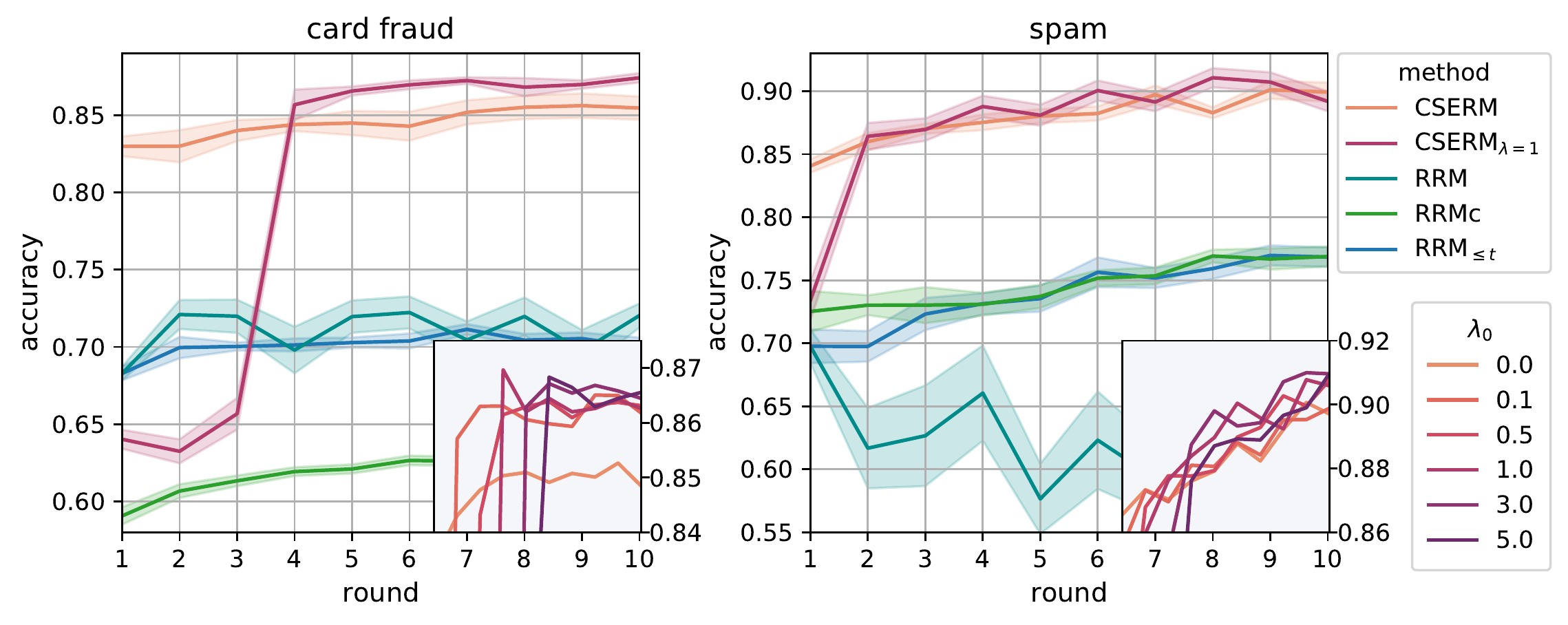}
    \includegraphics[width=0.35\textwidth]{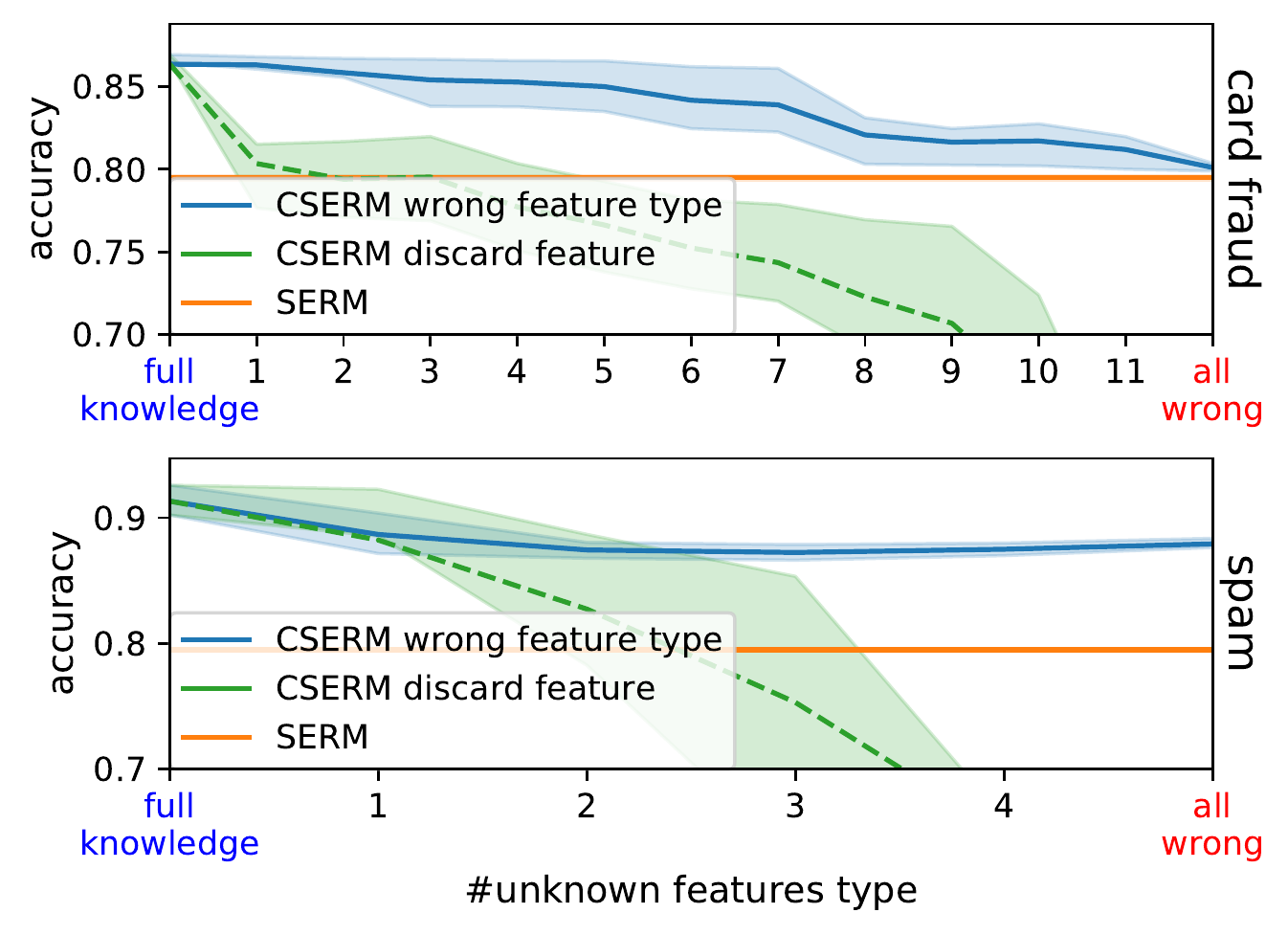}
    \caption{
    \textbf{(Left:)} Results for temporal methods over rounds.
    Inlay shows accuracy of \CSERMlambda\ for different values of $\lambda$.
    \textbf{(Right:)} Sensitivity analysis, showing
    accuracy for increasingly 'wrong' feature type attribution
    (causal vs. correlative).
    }
    \label{fig:time+sensitivity}
\end{figure*}

\paragraph{Methods.}
In addition to (i) \ERM, (ii) \SERM, and (iii) our \CSERM,
here we consider the following additional methods:
(iv) repeated risk minimization (\RRM) \citep{perdomo2020performative},
which applies \ERM\ independently at each round;
(v) \RRMlet, which uses all previous data;
and (vi) \RRMc, which uses only causal features to avoid dealing with `gaming' behavior.
For our approach, we distinguish between non-regularized (\CSERM)
and regularized (\CSERMreg{$\lambda$}) variants
(by default $\lambda_0=1$).
We include the non-strategic \bench,
and an \CSERMoracle-like 
benchmark
which combines $h^*$ within our approach.\looseness=-1





\subsection{Utilizing improvement on \texttt{card fraud}}
For \texttt{fraud}, we set $h^*$ be 3-layer MLP, on top of which we add noise,
so that some regions of $x$-space include a mixture of
positive and negative labels.
Potentially, if $f$ can incentivize such points
to move to areas where $h^*$ is more positive,
then this should entail better accuracy.
Table~\ref{tbl:real_experiments} (left) shows results.
As can be seen, \CSERM\ 
improves significantly over other methods,
gaining +8.2\% by accounting for changes in $y$ (vs. \SERM)
in addition to strategic effects (+19.9\% vs. \ERM).
which here \RRM\ is also able to achieve using time.
Regularization adds +1.2\%.
To gain insight as to why,
notice that \CSERM\ incentivizes more movement (14\% vs. \SERM);
of this, 14\% of points shift from $y=-1$ to $y^f=1$,
giving an overall improvement rate of 12\%.
In comparison, other methods improve by $<1\%$.
Improvement, however, does not imply that users necessarily benefit:
when considering welfare, defined as average utility minus costs
(and so in $[-1,1]$), the predictive success of \CSERM\ comes at the price of reduced welfare, \emph{despite} improvement.

\subsection{Avoiding pitfalls on \texttt{spam}}
For \texttt{spam}, we set $h^*$ to be linear,
except for one `tricky' causal feature $c_i$ which we concavify
by preserving its positive slope \emph{in-domain},
but reversing its slope to negative \emph{out-of-domain}.
This mimics a setting where having some amount of $c_i$ is helpful for obtaining $y=1$, but having `too much' is not.
Table~\ref{tbl:real_experiments} (right) shows results.
Here as well, \CSERM\ exhibits significant gains,
but this time through different means.
To see this, notice first  that \SERM's failure
comes from causing positive points to become negative (7.2\%);
this occurs since its reliance on $c_i$---which is predictively-useful in-domain---breaks once points move
out-of-domain in that direction
(interestingly, \SERM\ is blind to this, as its `perceived' accuracy
7\% higher than its actual accuracy).
Conversely, and by correctly identifying its nature,
\CSERM\ dodges the $c_i$ `trap', and diverts movement elsewhere.
\end{spacing}

\subsection{Time and regularization}
Fig. \ref{fig:time+sensitivity} (left) shows the performance of temporal methods over time.
Over time, and by utilizing additional dirty data,
\CSERM\ is able to improve performance 
(relative to $t=1$)
by ${\sim}3\%$ in \texttt{card fraud},  and ${\sim}6\%$ in \texttt{spam}.
\RRMc\ also improves over time---but to a significantly lesser degree; this shows the effectiveness of using causally-affected dirty data, 
but at the same time, reveals the (unutilized) potential of using non-causal features.
\RRM\ does use all features, but its performance over time is unstable
(in \texttt{spam} performance \emph{decreases} over time).
\RRMlet\ does improve, but is inconsistent across datasets.
As for the effect of regularization, results show how \CSERMreg{$\lambda$}\ initially performs worse than \CSERM---but proceeds to outperform it for both datasets. This holds for all $\lambda$, and becomes more pronounced (for better and worse) as $\lambda$ grows (see inlays).\looseness=-1


\subsection{Sensitivity analysis} \label{sec:exp_sensitivity}
Our final experiment tests the sensitivity of our approach to errors in features type attribution (i.e., considering a causal feature as non-causal, and vice versa).
Towards this, for each $d' \le d$,
we evaluate a variant of \CSERM\ which wrongly associates the type of a random subset of $d'$ features (\CSERMWRONG).
We compare this to \SERM\
(which does not use feature type information at all),
and to a variant of \CSERM\ which simply discards the wrong features (\CSERMDROPS).
Results are shown in Fig. \ref{fig:time+sensitivity} (right),
with average and standard deviation over 10 random seeds and $\min\{30, {d \choose d'}\}$ feature subsets per experimental condition.
As expected, errors in feature type attribution do entail reduced performance for \CSERMWRONG;
however, performance goes down slowly, either reaching \SERM\ when all features are wrong ($d'=d$; for \texttt{card fraud}), or remaining above it (\texttt{spam}).
In contrast, discarding the same `wrong' features causes performance to deteriorate quickly and sharply.

\section{Discussion}
This paper extends the study of strategic classification to causal settings in which changing inputs can also change outputs.
By focusing on the fundamental goal of optimizing accuracy,
our analysis surfaces the need for learning to accommodate two interwoven forms of distribution shift.
These differ in the challenges they present,
but are also complementary in their relation to \emph{time};
our approach utilizes these properties to provide a learning algorithm that is effective and efficient.
Our choice of remaining true to the original problem formulation
permits a clean formulation, and allows us to make connections to existing works.
Nonetheless, the current literature on strategic classification
remains far from being applicable in real social settings;
we view our work as taking one step toward this ultimate goal.


\subsection*{Acknowledgements}
This research was supported by the Israel Science Foundation (grant No. 278/22).

 
\bibliography{refs}

\begin{thebibliography}{35}
\providecommand{\natexlab}[1]{#1}
\providecommand{\url}[1]{\texttt{#1}}
\expandafter\ifx\csname urlstyle\endcsname\relax
  \providecommand{\doi}[1]{doi: #1}\else
  \providecommand{\doi}{doi: \begingroup \urlstyle{rm}\Url}\fi

\bibitem[Ahmadi et~al.(2021)Ahmadi, Beyhaghi, Blum, and
  Naggita]{ahmadi2021strategic}
Ahmadi, S., Beyhaghi, H., Blum, A., and Naggita, K.
\newblock The strategic perceptron.
\newblock In \emph{Proceedings of the 22nd ACM Conference on Economics and
  Computation}, pp.\  6--25, 2021.

\bibitem[Ahmadi et~al.(2022)Ahmadi, Beyhaghi, Blum, and
  Naggita]{ahmadi2022classification}
Ahmadi, S., Beyhaghi, H., Blum, A., and Naggita, K.
\newblock On classification of strategic agents who can both game and improve.
\newblock \emph{arXiv preprint arXiv:2203.00124}, 2022.

\bibitem[Alon et~al.(2020)Alon, Dobson, Procaccia, Talgam-Cohen, and
  Tucker-Foltz]{alon2020multiagent}
Alon, T., Dobson, M., Procaccia, A., Talgam-Cohen, I., and Tucker-Foltz, J.
\newblock Multiagent evaluation mechanisms.
\newblock In \emph{Proceedings of the AAAI Conference on Artificial
  Intelligence}, volume~34, pp.\  1774--1781, 2020.

\bibitem[Barsotti et~al.(2022)Barsotti, Ko{\c{c}}er, and
  Santos]{barsotti2022transparency}
Barsotti, F., Ko{\c{c}}er, R.~G., and Santos, F.~P.
\newblock Transparency, detection and imitation in strategic classification.
\newblock In \emph{Proceedings of the 31st International Joint Conference on
  Artificial Intelligence, IJCAI 2022}, 2022.

\bibitem[Bechavod et~al.(2021)Bechavod, Ligett, Wu, and
  Ziani]{bechavod2021gaming}
Bechavod, Y., Ligett, K., Wu, S., and Ziani, J.
\newblock Gaming helps! {L}earning from strategic interactions in natural
  dynamics.
\newblock In \emph{International Conference on Artificial Intelligence and
  Statistics}, pp.\  1234--1242. PMLR, 2021.

\bibitem[Bechavod et~al.(2022)Bechavod, Podimata, Wu, and
  Ziani]{bechavod2022information}
Bechavod, Y., Podimata, C., Wu, S., and Ziani, J.
\newblock Information discrepancy in strategic learning.
\newblock In \emph{International Conference on Machine Learning}, pp.\
  1691--1715. PMLR, 2022.

\bibitem[Br{\"u}ckner et~al.(2012)Br{\"u}ckner, Kanzow, and
  Scheffer]{bruckner2012static}
Br{\"u}ckner, M., Kanzow, C., and Scheffer, T.
\newblock Static prediction games for adversarial learning problems.
\newblock \emph{The Journal of Machine Learning Research}, 13\penalty0
  (1):\penalty0 2617--2654, 2012.

\bibitem[Chen et~al.(2020)Chen, Liu, and Podimata]{chen2020learning}
Chen, Y., Liu, Y., and Podimata, C.
\newblock Learning strategy-aware linear classifiers.
\newblock \emph{Advances in Neural Information Processing Systems},
  33:\penalty0 15265--15276, 2020.

\bibitem[Chen et~al.(2021)Chen, Wang, and Liu]{chen2021linear}
Chen, Y., Wang, J., and Liu, Y.
\newblock Linear classifiers that encourage constructive adaptation.
\newblock In \emph{Algorithmic Recourse workshop at ICML'21}, 2021.

\bibitem[Costa et~al.(2014)Costa, Merschmann, Barth, and
  Benevenuto]{costa2014pollution}
Costa, H., Merschmann, L.~H., Barth, F., and Benevenuto, F.
\newblock Pollution, bad-mouthing, and local marketing: the underground of
  location-based social networks.
\newblock \emph{Information Sciences}, 279:\penalty0 123--137, 2014.

\bibitem[Dong et~al.(2018)Dong, Roth, Schutzman, Waggoner, and
  Wu]{dong2018strategic}
Dong, J., Roth, A., Schutzman, Z., Waggoner, B., and Wu, Z.~S.
\newblock Strategic classification from revealed preferences.
\newblock In \emph{Proceedings of the 2018 ACM Conference on Economics and
  Computation}, pp.\  55--70, 2018.

\bibitem[Drusvyatskiy \& Xiao(2022)Drusvyatskiy and
  Xiao]{drusvyatskiy2022stochastic}
Drusvyatskiy, D. and Xiao, L.
\newblock Stochastic optimization with decision-dependent distributions.
\newblock \emph{Mathematics of Operations Research}, 2022.

\bibitem[Eilat et~al.(2022)Eilat, Finkelshtein, Baskin, and
  Rosenfeld]{eilat2022strategic}
Eilat, I., Finkelshtein, B., Baskin, C., and Rosenfeld, N.
\newblock Strategic classification with graph neural networks.
\newblock \emph{arXiv preprint arXiv:2205.15765}, 2022.

\bibitem[Estornell et~al.(2021)Estornell, Das, Liu, and
  Vorobeychik]{estornell2021unfairness}
Estornell, A., Das, S., Liu, Y., and Vorobeychik, Y.
\newblock Unfairness despite awareness: Group-fair classification with
  strategic agents.
\newblock In \emph{Thirty-fifth Conference on Neural Information Processing
  Systems (NeurIPS), StratML workshop}, 2021.

\bibitem[Ghalme et~al.(2021)Ghalme, Nair, Eilat, Talgam-Cohen, and
  Rosenfeld]{ghalme2021strategic}
Ghalme, G., Nair, V., Eilat, I., Talgam-Cohen, I., and Rosenfeld, N.
\newblock Strategic classification in the dark.
\newblock In \emph{International Conference on Machine Learning}, pp.\
  3672--3681. PMLR, 2021.

\bibitem[Haghtalab et~al.(2020)Haghtalab, Immorlica, Lucier, and
  Wang]{haghtalab2020maximizing}
Haghtalab, N., Immorlica, N., Lucier, B., and Wang, J.~Z.
\newblock Maximizing welfare with incentive-aware evaluation mechanisms.
\newblock In Bessiere, C. (ed.), \emph{Proceedings of the Twenty-Ninth
  International Joint Conference on Artificial Intelligence, {IJCAI-20}}, pp.\
  160--166, 7 2020.
\newblock Main track.

\bibitem[Hardt et~al.(2016)Hardt, Megiddo, Papadimitriou, and
  Wootters]{hardt2016strategic}
Hardt, M., Megiddo, N., Papadimitriou, C., and Wootters, M.
\newblock Strategic classification.
\newblock In \emph{Proceedings of the 2016 ACM conference on innovations in
  theoretical computer science}, pp.\  111--122, 2016.

\bibitem[Harris et~al.(2022)Harris, Ngo, Stapleton, Heidari, and
  Wu]{harris2022strategic}
Harris, K., Ngo, D. D.~T., Stapleton, L., Heidari, H., and Wu, S.
\newblock Strategic instrumental variable regression: Recovering causal
  relationships from strategic responses.
\newblock In \emph{International Conference on Machine Learning}, pp.\
  8502--8522. PMLR, 2022.

\bibitem[Jagadeesan et~al.(2021)Jagadeesan, Mendler-D{\"u}nner, and
  Hardt]{jagadeesan2021alternative}
Jagadeesan, M., Mendler-D{\"u}nner, C., and Hardt, M.
\newblock Alternative microfoundations for strategic classification.
\newblock In \emph{International Conference on Machine Learning}, pp.\
  4687--4697. PMLR, 2021.

\bibitem[Kleinberg \& Raghavan(2020)Kleinberg and
  Raghavan]{kleinberg2020classifiers}
Kleinberg, J. and Raghavan, M.
\newblock How do classifiers induce agents to invest effort strategically?
\newblock \emph{ACM Transactions on Economics and Computation (TEAC)},
  8\penalty0 (4):\penalty0 1--23, 2020.

\bibitem[Lechner \& Urner(2021)Lechner and Urner]{lechner2021learning}
Lechner, T. and Urner, R.
\newblock Learning losses for strategic classification.
\newblock In \emph{Thirty-fifth Conference on Neural Information Processing
  Systems (NeurIPS), Workshop on Learning in Presence of Strategic Behavior},
  2021.

\bibitem[Levanon \& Rosenfeld(2021)Levanon and Rosenfeld]{levanon2021strategic}
Levanon, S. and Rosenfeld, N.
\newblock Strategic classification made practical.
\newblock In \emph{International Conference on Machine Learning}, pp.\
  6243--6253. PMLR, 2021.

\bibitem[Levanon \& Rosenfeld(2022)Levanon and
  Rosenfeld]{levanon2022generalized}
Levanon, S. and Rosenfeld, N.
\newblock Generalized strategic classification and the case of aligned
  incentives.
\newblock In \emph{Proceedings of the 39th International Conference on Machine
  Learning (ICML)}, 2022.

\bibitem[Maheshwari et~al.(2022)Maheshwari, Chiu, Mazumdar, Sastry, and
  Ratliff]{maheshwari2022zeroth}
Maheshwari, C., Chiu, C.-Y., Mazumdar, E., Sastry, S., and Ratliff, L.
\newblock Zeroth-order methods for convex-concave min-max problems:
  Applications to decision-dependent risk minimization.
\newblock In \emph{International Conference on Artificial Intelligence and
  Statistics}, pp.\  6702--6734. PMLR, 2022.

\bibitem[Mendler-D{\"u}nner et~al.(2022)Mendler-D{\"u}nner, Ding, and
  Wang]{mendler2022predicting}
Mendler-D{\"u}nner, C., Ding, F., and Wang, Y.
\newblock Predicting from predictions.
\newblock In \emph{Advances in neural information processing systems}, 2022.

\bibitem[Miller et~al.(2020)Miller, Milli, and Hardt]{miller2020strategic}
Miller, J., Milli, S., and Hardt, M.
\newblock Strategic classification is causal modeling in disguise.
\newblock In \emph{International Conference on Machine Learning}, pp.\
  6917--6926. PMLR, 2020.

\bibitem[Miller et~al.(2021)Miller, Perdomo, and Zrnic]{miller2021outside}
Miller, J.~P., Perdomo, J.~C., and Zrnic, T.
\newblock Outside the echo chamber: Optimizing the performative risk.
\newblock In \emph{International Conference on Machine Learning}, pp.\
  7710--7720. PMLR, 2021.

\bibitem[Nair et~al.(2022)Nair, Ghalme, Talgam-Cohen, and
  Rosenfeld]{nair2022strategic}
Nair, V., Ghalme, G., Talgam-Cohen, I., and Rosenfeld, N.
\newblock Strategic representation.
\newblock In \emph{International Conference on Machine Learning}, pp.\
  16331--16352. PMLR, 2022.

\bibitem[Perdomo et~al.(2020)Perdomo, Zrnic, Mendler-D{\"u}nner, and
  Hardt]{perdomo2020performative}
Perdomo, J., Zrnic, T., Mendler-D{\"u}nner, C., and Hardt, M.
\newblock Performative prediction.
\newblock In \emph{International Conference on Machine Learning}, pp.\
  7599--7609. PMLR, 2020.

\bibitem[Shavit et~al.(2020)Shavit, Edelman, and Axelrod]{shavit2020causal}
Shavit, Y., Edelman, B., and Axelrod, B.
\newblock Causal strategic linear regression.
\newblock In \emph{International Conference on Machine Learning}, pp.\
  8676--8686. PMLR, 2020.

\bibitem[Shimodaira(2000)]{shimodaira2000improving}
Shimodaira, H.
\newblock Improving predictive inference under covariate shift by weighting the
  log-likelihood function.
\newblock \emph{Journal of statistical planning and inference}, 90\penalty0
  (2):\penalty0 227--244, 2000.

\bibitem[Sundaram et~al.(2021)Sundaram, Vullikanti, Xu, and
  Yao]{sundaram2021pac}
Sundaram, R., Vullikanti, A., Xu, H., and Yao, F.
\newblock {PAC}-learning for strategic classification.
\newblock In \emph{International Conference on Machine Learning}, pp.\
  9978--9988. PMLR, 2021.

\bibitem[Tsirtsis \& Gomez~Rodriguez(2020)Tsirtsis and
  Gomez~Rodriguez]{tsirtsis2020decisions}
Tsirtsis, S. and Gomez~Rodriguez, M.
\newblock Decisions, counterfactual explanations and strategic behavior.
\newblock \emph{Advances in Neural Information Processing Systems},
  33:\penalty0 16749--16760, 2020.

\bibitem[Zhang \& Conitzer(2021)Zhang and Conitzer]{zhang2021incentive}
Zhang, H. and Conitzer, V.
\newblock Incentive-aware {PAC} learning.
\newblock In \emph{Proceedings of the AAAI Conference on Artificial
  Intelligence}, volume~35, pp.\  5797--5804, 2021.

\bibitem[Zrnic et~al.(2021)Zrnic, Mazumdar, Sastry, and Jordan]{zrnic2021leads}
Zrnic, T., Mazumdar, E., Sastry, S., and Jordan, M.
\newblock Who leads and who follows in strategic classification?
\newblock \emph{Advances in Neural Information Processing Systems}, 34, 2021.

\end{thebibliography}
\bibliographystyle{icml2023}

\newpage
\appendix
\onecolumn

\section{Proofs} \label{apx:proofs}
\todo{add other things we need proofs for}

\subsection{Lemma \ref{lem:only_xr}}\label{proof_lem_1}
\begin{proof}

Let $p(x_r,y)$ be some base distribution, and let $f$ be a classifier $f: \mathcal{X}_{r} \rightarrow \mathcal{Y}$. 
Given $f$, denote $x_r'=\Delta_f(x_r)$,
for which we can write the induced joint distribution as $p^f(x_r' y')$.
To consider both $x_r$ and $x'_r$ together, we denote
their joint distribution with $y$ by $q^f(x_r,x_r',y')$, whose definition derives immediately from $\Delta_f$.
Since $\Delta_f$ is deterministic, we get that $q^f(x_r'|x_r)=\ind{\Delta_f(x_r)=x_r'}=\ind{x_{r} \in \Delta_{f}^{\minus 1}(x_{r}')}$.

First, with the law of total probability, we get the following expression for the induced marginal density:
\begin{equation}\label{eq:x_r marginal proof}
    p^f(x_r') = \int_{x_r\in \mathcal{X}_{r}}p(x_r)q^f(x_r'|x_r)\d x_r = \int_{x_r\in\mathcal{X}_{r}}p(x_r)\cdot\ind{x_{r} \in \Delta_{f}^{\minus 1}(x_{r}')}\d x_r = \int_{x_{r} \in \Delta_{f}^{\minus 1}(x_{r}')}
p(x_{r}) \d x_r
\end{equation}
For the induced conditional density, again using the law of total probability we get:
\begin{equation}
    p^f(y|x_r') = \int_{x_r\in \mathcal{X}_{r}} q^f(x_r|x_r')q^f(y|x_r', x_r)\d x_r
\end{equation}
Now, since $y$ is sampled jointly with $x_r$ from $p$, and since $x_r'$ is a function only of $x_r$ (which in itself is non-causal), we get that $q^f(y|x_r,x_r')=p(y|x_r)$.
Also, from Bayes' theorem, we have $q^f(x_r|x_r')=q^f(x'_r|x_r)\frac{p(x_r)}{p^f(x'_r)}$.
With these we get 
\begin{equation}
\begin{aligned}
    p^f(y|x_r')
    &= \int_{x_r\in \mathcal{X}_{r}} q^f(x'_r|x_r)\frac{p(x_r)}{p^f(x'_r)}p(y|x_r)\d x_r \\
    &=
    \int_{x_r\in \mathcal{X}_{r}} \ind{x_{r} \in \Delta_{f}^{\minus 1}(x_{r}')}\cdot\frac{p(x_r)}{p^f(x'_r)}p(y|x_r)\d x_r \\
    &= \int_{x_{r} \in \Delta_{f}^{\minus 1}(x_{r}')}\frac{p(x_r)}{p^f(x'_r)}p(y|x_r)\d x_r  \
\end{aligned}
\end{equation}
\end{proof}

\subsection{Lemma \ref{lem:xc_and_xr}}\label{proof_lem_3}
\begin{proof}
Let $p(x, u, y)$ be some base distribution, and let $f$ be a classifier $f: \mathcal{X} \rightarrow \mathcal{Y}$. 
Given $f$, we denote $x'=\Delta_f(x)$, and define the joint distributions $p^f(x', u, y')$ and $q^f(x,u,x',y')$.
From Eq. \eqref{eq:x_r marginal proof} and replacing $x_r$ with $x$, we get:
\begin{equation}
    p^f(x') = \int_{x \in \Delta_{f}^{\minus 1}(x')}
    p(x) \d x
\end{equation}
For the induced conditional density, with the law of total probability, we get:
\begin{equation}
    p^f(y'|x')= \int_{u\in \mathcal{U}} p^f(u|x')p^f(y'|x',u)\d u
\end{equation}
For a stochastic $h^*=p^*$, since $x_c,u$ are both causal (and $x_r$ is not),
and since they jointly fully determine $y$ (up to irreducible noise in $h^*$),
we get that:
\[
p^f(y'|x',u)=p^f(y'|x_c',x_r',u)=p^f(y'|x_c',u)=p^*(y'|x_c',u)
\]
Plugging in we get:
\begin{equation}\label{eq:x cond}
    p^f(y'|x')= \int_{u\in \mathcal{U}} p^f(u|x')p^*(y'|x_c',u)\d u
\end{equation}

Again using the law of total probability, generally we have:
\begin{equation}
    p^f(u|x')= \int_{x\in\mathcal{X}} q^f(x|x') q^f(u|x', x)\d x
\end{equation}
However, this can be simplified using the fact that $x'$ originates from some $x$,
i.e., $x' = \Delta_f(x)$.
For the second term, 
since $x'$ is a function of $x$ alone, we get that $q^f(u|x', x) = p(u|x)$.
For the first term, and similarly to the proof in \ref{proof_lem_1}, using Bayes' theorem and the definition of $\Delta_{f}^{\minus 1}$ we get:
\[
q^f(x|x')=q^f(x'|x)\frac{p(x)}{p^f(x')} = \ind{x \in \Delta_{f}^{\minus 1}(x')}\cdot\frac{p(x)}{p^f(x')}
\]
Plugging into the equation above gives:
\begin{equation}
\begin{aligned}
    p^f(u|x')&= \int_{x\in\mathcal{X}} \ind{x \in \Delta_{f}^{\minus 1}(x')}\cdot\frac{p(x)}{p^f(x')} p(u|x)\d x \\
    &= \int_{x\in\Delta_{f}^{\minus 1}(x')}\frac{p(x)}{p^f(x')} p(u|x)\d x \\
    &= \int_{x\in\Delta_{f}^{\minus 1}(x')}\frac{p(x|u)}{p^f(x')} p(u)\d x
\end{aligned}
\end{equation}

With the definition of $\nu_f$ as:
\begin{equation}
    \nu_f(u;x') = 
    \int_{x \in \Delta_{f}^{\minus 1}(x')}
    \frac{p(x | u)}{p^f(x')} \d x
\end{equation}
Taking out $p(u)$ we can write:
\begin{equation}
    p^f(u|x') = \nu_f(u;x') p(u) 
\end{equation}
Plug it in back to Eq. \eqref{eq:x cond}, we get:
\begin{equation}
    p^f(y'|x')= \int_{u\in \mathcal{U}} \nu_f(u;x') p(u) p^*(y'|x_c',u)\d u
\end{equation}

In the case of a deterministic $h^*$, i.e. $p^*(y'|x_c', u)= \ind{h^*(x_c',u) = y'}$, this simplifies to:
\begin{equation}
\begin{aligned}
    p^f(y'|x')&= 
    \int_{u\in \mathcal{U}} \nu_f(u;x') p(u)\cdot \ind{h^*(x_c',u) = y'}\d u \\
    &= \int_{u:\:y'= h^*(x'_{c}, u)} \nu_f(u;x') p(u) \d u
\end{aligned}
\end{equation}
Additionally, in the case where $x,u$ are independent, we get that $p(x|u)=p(x)$, therefore:
\begin{align}
    \nu_f(u;x') &= 
    \int_{x \in \Delta_{f}^{\minus 1}(x')}
    \frac{p(x | u)}{p^f(x')} \d x \\
    & = \int_{x \in \Delta_{f}^{\minus 1}(x')}
    \frac{p(x)}{p^f(x')} \d x = 1 \nonumber
\end{align}

which gives:
\begin{equation}\label{eq:x and u indep}
    p^f(y'|x')= 
    \int_{u:\:y'= h^*(x'_{c}, u)} p(u) \d u
\end{equation}
\end{proof}

\subsection{Lemma \ref{lem:only_xc}}
\begin{proof}
This is a special case of Lemma \ref{lem:xc_and_xr}. Replacing $x$ with $x_c$ in Eq. \eqref{eq:x and u indep} completes the proof.
\end{proof}

\section{Additional results}
\subsection{Accuracy and improvement can be at odds}\label{apx:acc_vs_improve}

\begin{figure}[h]
    \centering
    \includegraphics[width=0.3\columnwidth]{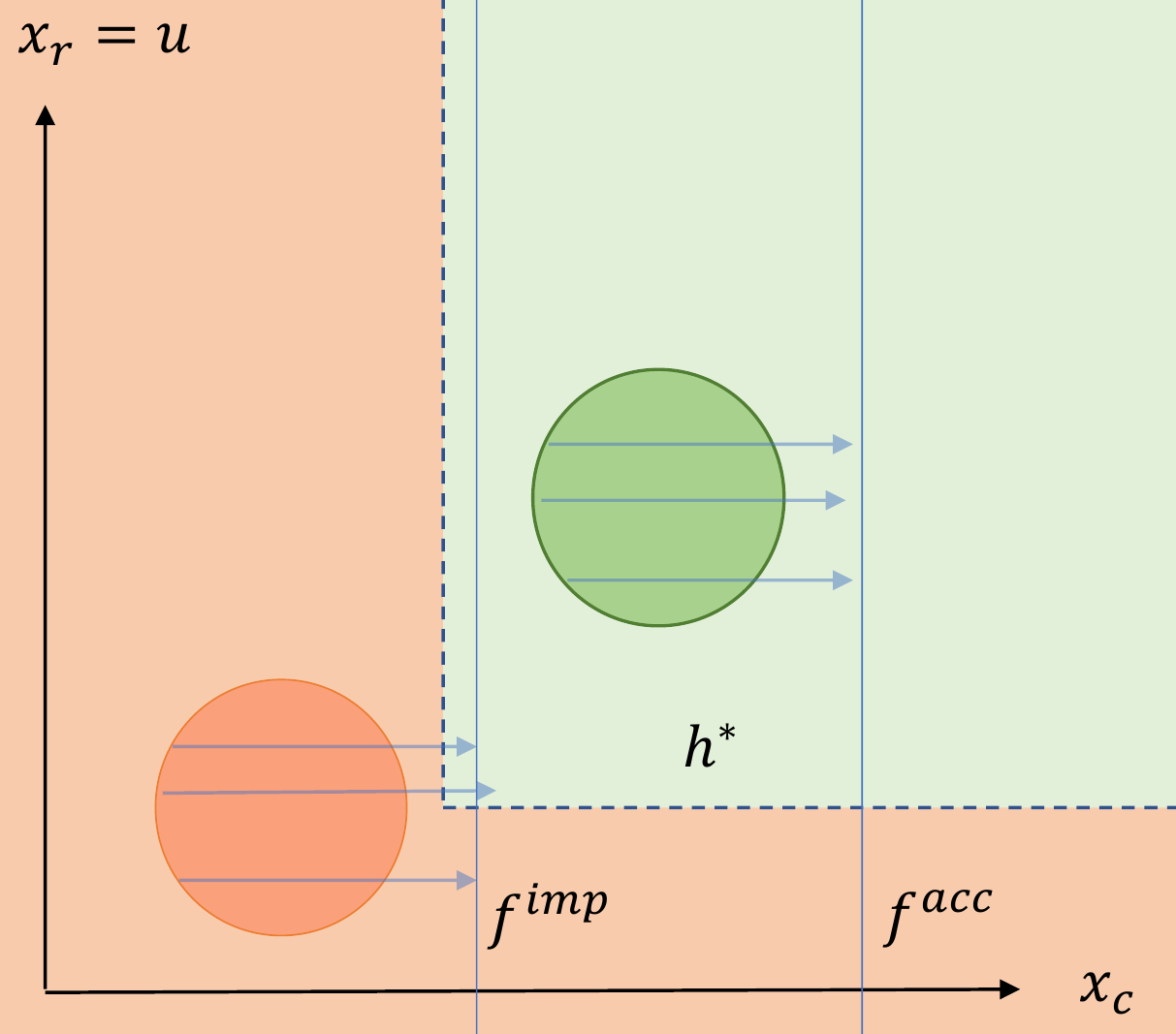}
    \caption{
    An optimal classifier for improvement is not necessarily optimal for accuracy.}
    \label{fig:improvement is bad}
\end{figure}

Fig. \ref{fig:improvement is bad} illustrates the idea that an optimal classifier in terms of maximizing improvement is not necessarily optimal for maximizing accuracy.
In this example, the red and the green circles represent clusters of negative points ($y=-1$) and positive points ($y=1$) respectively. 
The decision boundary of $h^*$ is illustrated by a dashed line, and $x_r=u$.
$f^{\text{imp}}$ makes all the negative points move from the red circle to its decision boundary; half of the points (the upper half of the circle) become positive ($y'=1$) since after movement their projection on their original $u=x_r$ lies in the positive region of $h^*$, and half of the points (the lower half of the circle) remains negative ($y'=-1$) since after movement their projection on their original $x_r$ lies in the negative region of $h^*$.
The points from the lower half of the red circle could never become positive: no matter how they move, their projection on their original $x_r$ will always lie in the negative region of $h^*$. 
Therefore, $f^{\text{imp}}$ turns all the possibly improvable points into positive and keeps all the originally positive points positive, hence it is optimal for maximizing improvement. 
However, since the points from the lower half of the red circle move to the decision boundary of $f^{\text{imp}}$, they are classified as positive ($\yhat=1$), which means $f^{\text{imp}}$ err ($\yhat \neq y'$) on each point from the lower half of the red circle.
In contrast, $f^{\text{acc}}$ make only the positive points from the green circle to move, and after moving their projection on their original $x_r$ stays in the positive region of $h^*$, therefore they are classified correctly ($\yhat=y'=1$); since the negative points from the red cluster don't move they are also classified correctly ($\yhat=y'=-1$), which means $f^{\text{acc}}$ gets $100\%$ accuracy, hence it is an optimal classifier for maximizing accuracy, with higher accuracy than $f^{\text{imp}}$.

\subsection{Efficient computation of $\xtilde_r$} \label{apx:efficient_computation}
\guy{maybe we will drop the word 'efficient'? I'm not sure how much my computation is efficient}
In this section, we show how to efficiently compute $\xtilde_r = \expect{x_r \sim \phat(x_r | x^f)}{x_r}$ for a linear $f$ and a generalized quadratic cost
$c_Q(x,x') = (x'-x)^\top Q (x'-x) = \|x'-x\|_Q^2$
for PSD $Q$.
Recall that the idea underlying our definition of $\xtilde_r$
is that we'd like to `reconstruct' $x_r$, to the best of our ability, given a strategically modified example $x^f$.
This is done by considering its likelihood, $p^f(x_r|x^f)$.
We can express this likelihood using the clean marginal density:
\begin{equation}
\begin{aligned}
    p^f(x_r|x^f)&= \frac{p(x_r)}{p^f(x^f)}p^f(x^f|x_r)= \frac{p(x_r)}{p^f(x^f)}\int_{x_c} p(x_c|x_r)p^f(x^f|x_c,x_r)\d x_c \\
    &= \frac{p(x_r)}{p^f(x^f)}\int_{x_c\in \mathcal{X}_c} p(x_c|x_r)\cdot\ind{(x_c,x_r)\in\Delta_f^{\minus 1}(x^f)}\d x_c \\
    &= \frac{p(x_r)}{p^f(x^f)}\int_{x_c: \:(x_c,x_r)\in \Delta_f^{\minus 1}(x^f)} p(x_c|x_r)\d x_c\\
    &=  \frac{1}{p^f(x^f)}\int_{x_c: (x_c,x_r)\in \Delta_f^{-1}(x^f)} p(x_c,x_r)\d x_c
\end{aligned}
\end{equation}
where 
\begin{equation}
    p^f(x^f) =  \int_{x \in \Delta_{f}^{\minus 1}(x^f)}p(x)\d x
\end{equation}
Using a model of the clean marginal density $\phat(x) \approx p(x)$ we can estimate $p^f(x_r|x^f)$ for any point by replacing $p$ with $\phat$ in this expression.
The expected value of this likelihood is 
\begin{equation}\label{eq:xtilde exp}
\begin{aligned}
    \xtilde_r &= \expect{x_r \sim \phat(x_r | x^f)}{x_r} = \int_{x_r\in\mathcal{X}_r} \phat(x_r | x^f)\cdot x_r \d x_r \\
    &= \frac{1}{\phat^f(x^f)}  \int_{x_r\in\mathcal{X}_r} \int_{x_c: (x_c,x_r)\in \Delta_f^{-1}(x^f)}\phat(x_c, x_r)\cdot x_r \d x_c\d x_r \\
    &= \frac{1}{\phat^f(x^f)}  \int_{x\in \Delta_f^{-1}(x^f)}\phat(x)\cdot x_r \d x \\
\end{aligned}
\end{equation}

Next, we describe the precise structure of $\Delta_f^{-1}$
(for a linear classifier $f(x)=\sign (w^\top x + b)$ and $ \|x'-x\|_Q^2$ cost)
and show how it permits tractable computation.
In our setting, points move directly to the hyperplane, in a straight line, defined by $w$ and $b$, i.e., over a line that is orthogonal to the hyperplane.
This means that a point $x$ moves to $x^f= x - z\wbar$ such that $w^\top x^f = -b$, where $\wbar=\frac{w}{|w|}$ and $z\in\mathbb{R}$ is the movement step.
To see which points can afford to move, we can look at the "furthest" points from the hyperplane that can still afford the movement, i.e., points $x$ such that after movement to $x^f$ pay cost of $ \|x^f-x\|_Q^2 =2$. 
Since $Q$ is PSD, there is an invertible matrix $A$ such that $Q=A^\top A$, so we can rewrite the cost as $\|A(x^f-x)\|_2^2=\|Ax^f-Ax\|_2^2$. 
Therefore, the points that pay cost of $2$ are points such that $2=\|Ax^f- Ax\|_2^2=\|A(x - z\wbar) - Ax\|_2^2= 
\|Ax - zA\wbar - Ax\|_2^2= \|-z A\wbar\|_2^2=z^2\| A\bar{w}\|_2^2$. With this, we can get the maximal movement step that points can afford to do: $z=\frac{\sqrt{2}}{\| A\wbar\|_2}$.
Therefore, for a point $x^f$ which lies on the hyperplane, i.e. $w^\top x^f=-b$, we get that $\Delta_{f}^{-1}(x^f)=\left\{ x^f-z \bar{w}\Big| 0\leq z\leq \frac{\sqrt{2}}{\| A\bar{w}\|_2}\right\}$.
Plugging in this to Eq. \eqref{eq:xtilde exp}, we get:
\begin{align}
    \xtilde_r &=\frac{\int_{x\in \Delta_f^{-1}(x^f)}\phat(x)\cdot x_r \d x}{\int_{x'\in \Delta_f^{-1}(x^f)}\phat(x') \d x'}  \\
    & = 
    x_{r}^f-w_{r} \cdot \frac{\int_{z=0}^{\frac{\sqrt{2}}{\|A \bar{w}\|_2}}z \cdot \phat(x^f-z\bar{w}) \d z}{\int_{z'=0}^{\frac{\sqrt{2}}{\|A \bar{w}\|_2}} \phat(x^f-z' \bar{w}) \d z'} \nonumber
\end{align}
In practice, we compute these integrals numerically.

\section{Experimental details - synthetic data}
In all of our synthetic experiments, we used 500 samples for clean training data, 150 samples of dirty data collected at each round (out of total $T=10$ rounds), 100 samples for validation set, and 400 samples for test set.
We will now specify experimental details for each experiment.
\subsection{Experiment A - utilizing improvement}
\begin{enumerate}
    \item Structure of $h^*$: in this experiment, $h^*$ is a linear function wrapped with a stochastic mechanism that creates noisy labels near the decision boundary.
    \item Structure $p(x_c,u)$: $p(x_c,u)$ is constructed from 3 normal distributed clusters: i) cluster of positive points with $\mu=(2,2),\sigma^2=0.4$ that contains $15\%$ of the total points, ii) cluster of negative points with $\mu=(-5.5,-5.5),\sigma^2=0.6$ that contains $10\%$ of the total points, and iii) cluster of a mixture of positive and negative points with $\mu=(-2,-2),\sigma^2=0.6$ that contains $75\%$ of the total points.
    \item Cost scale = 0.035.
    \item Class of $h$: polynomial model with a degree of 3.
    \item Hyper-parameters: $f$ learning-rate = 0.01, $h$ learning-rate = 0.01, batch-size = 64, sigmoid temperature =4 , $l_2$ regularization coefficient for $f$ = 0, $l_2$ regularization coefficient for $h$ = 0.
\end{enumerate}

\subsection{Experiment B - avoiding pitfalls}
\begin{enumerate}
    \item Structure of $h^*$: in this experiment, $h^*$ has a circle shape with a center in $(0,0)$ where points inside the circle are labeled as positive and points outside it are labeled as negative.
    \item Structure $p(x_c,u)$: $p(x_c,u)$ is constructed from 2 normal distributed clusters: i) cluster of positive points with $\mu=(0,0),\sigma^2=0.3$ that contains $50\%$ of the total points, ii) cluster of negative points with $\mu=(-5.5,-5.5),\sigma^2=(0.3,0.45)$ that contains $50\%$ of the total points.
    \item Cost scale = 0.07
    \item Class of $h$: polynomial model with a degree of 3.
    \item Hyper-parameters: $f$ learning-rate = 0.1, $h$ learning-rate = 0.1, batch-size = 64, sigmoid temperature = 20, $l_2$ regularization coefficient for $f$ = 0.1, $l_2$ regularization coefficient for $h$ = 0.
\end{enumerate}

\subsection{Experiment C - XOR}
\begin{enumerate}
    \item Structure of $h^*$: in this experiment, $h^*$ is constructed from 3 ellipses: 2 vertical ellipses with centers $(2,0)$ and $(-2,0)$ and one horizontal ellipse with a canter $(0,0)$. Together these ellipses create a shape where points inside it are labeled as negative, and points outside it are labeled as negative.
    \item Structure $p(x_c,u)$: $p(x_c,u)$ is constructed from 4 normal distributed clusters, each contains $25\%$ of the points and with $\sigma^2=0.3$: i) cluster of positive points with $\mu=(0,2.5)$, ii) cluster of positive points with $\mu=(0,-2.5)$ iii) cluster of negative points with $\mu=(2.5,0)$, and iiii) cluster of negative points with $\mu=(-2.5,0)$.
    \item Cost scale = 0.08
    \item Class of $h$: polynomial model with a degree of 4.
    \item Hyper-parameters: $f$ learning-rate = 0.05, $h$ learning-rate = 0.01, batch-size = 64, sigmoid temperature = 20, $l_2$ regularization coefficient for $f$ = 1, $l_2$ regularization coefficient for $h$ = 0.01, exploration regularization coefficient $\lambda=5$ with decay of 0.4.
\end{enumerate}

\section{Experimental details - real data}
\label{apx:experiments_real}

\subsection{Data and preprocessing}
\subsubsection{card fraud}

\paragraph{Data description.}
The data is publicly available at \url{https://www.kaggle.com/datasets/mlg-ulb/creditcardfraud}.
This dataset contains transactions made by credit
cards that occurred in two days during September 2013 by European cardholders.
This data set is highly unbalanced and contained 492 frauds out of 284,807 transactions.
The data contains 31 numerical features: 'Time' which contains the seconds elapsed between each transaction and the first transaction in the dataset, 'Amount' which is the transaction amount, and additional 29 features which are the result of a PCA transformation.

\paragraph{Preprocessing.}
As preprocessing, we removed the 'Time' feature and then performed Z-score normalization to the data, followed by a division by the square root of the data dimension. 

\paragraph{Data augmentation.}
Since this dataset contains only 492 negative samples, we created synthetic negative samples for the experiment by fitting a KDE model to the negative samples and then sampling generated samples from the model.

\paragraph{Data split.}
We sampled 5500 balanced samples for the experiment and set 3000 of them ($\sim 54\%$) as training data, 500 ($\sim 9\%$) as validation data, and 2000 ($\sim 36\%$) as test data.
The baseline that doesn't use time used all of the training data in a single round. 
For the methods that do use time, including ours, we split the training data into 1000 clean samples and 2000 assigned to be dirty samples, partitioned into 10 batches of 200 samples each.
in the first round, they got access only to the clean samples, and then during 10 rounds, each round $t$, they got access to additional 200 dirty samples that were created by applying $\Delta_f$ on the $t$-batch of the dirty samples inventory.

\paragraph{Experiment repetition.}
We repeated the experiment 15 times, each time with a random data split. The reported results are the averages and standard error over these random splits.

\subsubsection{spam}
\paragraph{Data description.}
The data can be obtained by the authors of \citet{costa2014pollution}.
The data includes features describing users of a large social network, some of which are spammers.
The data is balanced with a total of 7076 samples and contains 60 numerical features and binary labels (spammer or not). 

\paragraph{Preprocessing.}
As preprocessing, we kept only 15 features: qTips\_plc, rating\_plc, qEmail\_tip, qContacts\_tip, qURL\_tip, qPhone\_tip, qNumeriChar\_tip, sentistrength\_tip, combined\_tip, qWords\_tip,             followers\_followees\_gph, qUnigram\_avg\_tip', qTips\_usr, indeg\_gph, qCapitalChar\_tip.
After removing the other features we performed Z-score normalization on the data, followed by a division by the square root of the data dimension. 

\paragraph{Data split.}
Same as in \texttt{card fraud}.

\paragraph{Experiment repetition.}
Same as in \texttt{card fraud}.



\subsection{Feature partition and labeling function $h^*$}

\subsubsection{card fraud}
\paragraph{Feature partition.}
After preprocessing we selected 6 features to be $x_c$, and 16 features to be $u$. We create $x_r$ by takings the first 6 features from $u$ and multiplying them by a random square matrix. 

\paragraph{Labeling function.}
We created $h^*$ by fitting an MLP with 3 hidden layers with hidden dimensions of 10 on a balanced subset of the original data (before augmenting it with a KDE). 
We then wrapped the MLP model with a stochastic mechanism that given an input $x$, assigns a probability $p$ as a function of the distance of $x$ from the decision boundary of the model and then multiply the scores MLP$(x)$ by $-1$ with a probability of $p$.
After assigning score $s$ to a sample, its label is $\sign(s)$.
In this way, points from the region near the decision boundary of $h^*$ have noisy labels and they are a mixture of negative and positive points.

\subsubsection{spam}
\paragraph{Feature partition.}
After preprocessing we selected 3 features to be $x_c$, and 12 features to be $u$. We create $x_r$ by takings the first 2 features from $u$ and multiplying them by a random square matrix. 

\paragraph{Labeling function.}
We created $h^*$ by first fitting a linear model $g$ on the data.
We then wrapped the $g$ with a 'tricky-feature' mechanism defined for a feature $c_i$, a threshold $\gamma$ and a slope $\beta$: given an input $x$, if $x_{c_i}>\gamma$, then replace the score $s=g(x)$ of the linear model with $s\gets g(x)-\beta(x_{c_i}-\gamma)$.
After assigning score $s$ to a sample, its label is $\sign(s)$.
We used $\gamma=0.05$ and $\beta=20$; these values were chosen such that this mechanism will cause label flip to only $\sim 5\%$ of the original data.

\subsection{Density estimation}
For both usages of a density model in our algorithm, $\phat$ and $\qhat$, we used KDE with a Gaussian kernel. 
The hyper-parameter of the model is the kernel bandwidth, which we choose using a grid search cross-validation.


\subsection{Training, tuning, and optimization}
For both \texttt{card fraud} and \texttt{spam} experiments, we used the following parameters, which we choose manually:
\begin{enumerate}
    \item class of $h$: MLP with 3 layers with a width of 10
    \item $f$, $h$ learning-rate = 0.01
    \item batch size = 64
    \item epochs = 100
    \item an early stopping mechanism when there are 7 consecutive epochs without accuracy improvement on the validation set
    \item sigmoid temperature $\tau=4$
    \item exploration regularization coefficients: in \CSERMreg{$\lambda=0.1$} we used $\lambda_0=0.1$ which decays in each round with factor of 0.4.
    in \CSERMreg{$\lambda=1$} we used $\lambda_0=1$ which decays in each round with factor of 0.4.
    
\end{enumerate}
In each experiment, we used a different cost scale $\alpha$, chosen such that there will $\sim 50\%$ of strategically moving points: at \texttt{card fraud} we used $\alpha=1$, and in \texttt{spam} we used $\alpha=40$.


\subsection{Baselines and benchmarks}
In our experiments, we used two benchmarks:
\begin{enumerate}
    \item \bench: the result of a \naive\ \ERM\ tested on a non-strategic test; this benchmark shows us the maximal possible accuracy when there is no strategic behavior.
    \item \CSERMoracle: our method (\CSERM) with oracle access to $h^*$ and $u$, therefore in training it can accurately fix $y\mapsto y'$, for moving points; this benchmark shows us the maximal possible accuracy in a causal strategic setting, where there are no information gaps to the learner.
\end{enumerate}
Additionally, we used the following baselines:
\begin{enumerate}
    \item \ERM: simulate a \naive\ learner who doesn't aware to strategic behaviour. 
    The results of this baseline show us how much the learner can lose by not accounting for strategic behavior.
    \item \SERM: a strategically-aware but causally-oblivious baseline that optimizes Eq. \eqref{eq:sc_learning_objective} using the strategic hinge loss \citep{levanon2022generalized}. The results of this baseline show us how much the learner can lose by accounting only for the strategic movement of $x$ and not for the possible change in the label $y$.
    \item \RRM: a baseline that uses time by collecting dirty data at each round, and at each round applies \ERM\ using only the last collected dataset.
    This baseline simulates a learner that is aware of the distribution shift, but either doesn't know the structure of the shift or simply doesn't know how to tackle the problem of the specific distribution shift caused by strategic behavior and causality.
    \item \RRMlet: a version of \RRM\ that at each round uses the collected data from all previous rounds. 
    This baseline simulates a learner that is aware of the fact that data from various distributions can be useful for learning under the distribution shift.
    \item \RRMc: a version of \RRMlet\ that uses only causal features and uses all previous data.
    This baseline simulates a learner that is aware of the causal strategic structure of the distribution shift, knows the partition of features to $x_c$ and $x_r$, and chooses to use only $x_c$ to avoid dealing with `gaming' behavior that the use of $x_r$ causes.
    \item \CSERM: our approach, without regularizing for exploration.
    \item \CSERMreg{$\lambda=0.1$}: our approach with exploration regularization coefficient of $0.1$ in the first round, and decaying with a factor of $0.4$ in each round.
    \item \CSERMreg{$\lambda=1$}: our approach with exploration regularization coefficient of $1$ in the first round, and decaying with a factor of $0.4$ in each round.
\end{enumerate}

\begin{figure}[t!]
    \centering
    \includegraphics[width=0.6\textwidth]{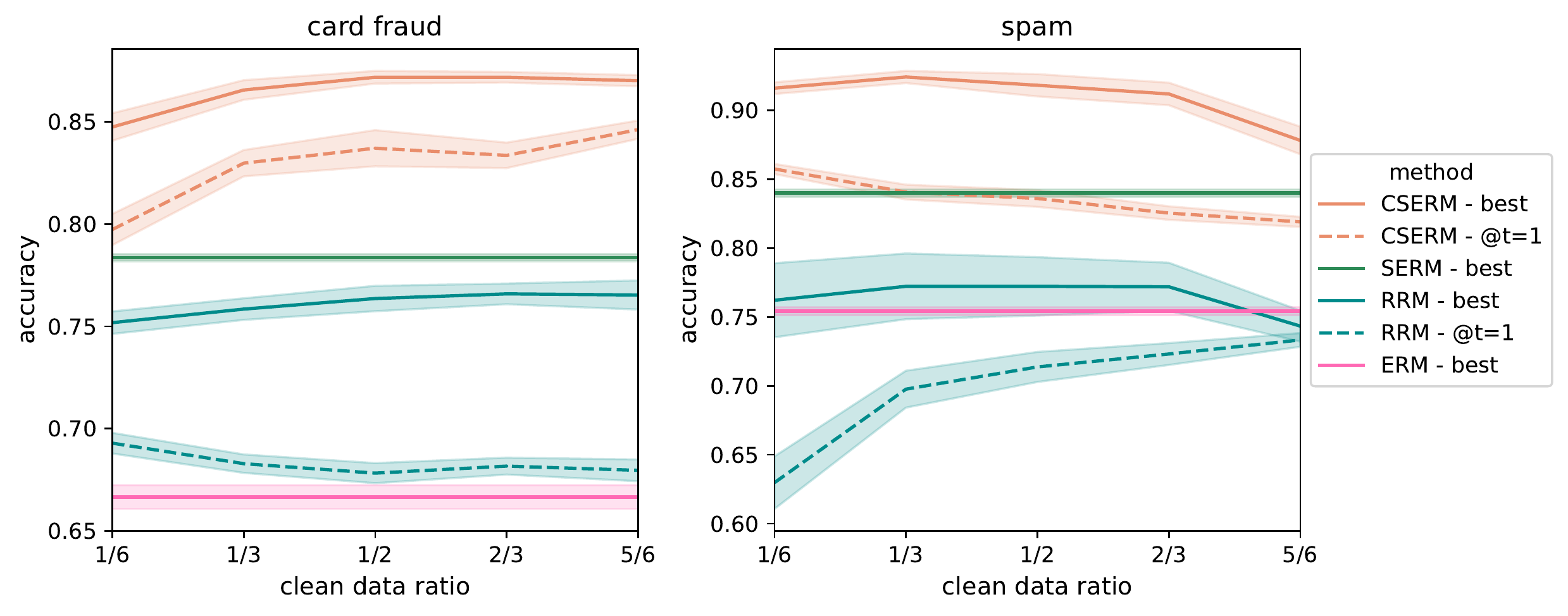}
    \caption{
    Accuracy across different ratios of clean vs dirty data.
    }    \label{fig:changing_clean_data_ratio}
\end{figure}

\section{Additional experimental results}
\label{apx:experiments_extra}

\subsection{Varying clean data ratio}\label{apx:experiments_clean_data_ratio}
This experiment tests the effect of the ratio of clean vs. dirty data on the performance of temporal methods that use dirty data over time in addition to clean data.
Towards this, for each $r\in \{\frac{1}{6}, \frac{1}{3}, \frac{1}{2}, \frac{2}{3}, \frac{5}{6}\}$ we assigned an $r$-fraction of the training data to include clean example, and the remaining $1-r$-fraction to include dirty samples, while keeping the total size of training data fixed to 3,000 samples. 
Figure \ref{fig:changing_clean_data_ratio} plots performance as a function of $r$.
As can be seen, the overall trend of the effect of $r$ on accuracy changes across methods and datasets.
However, results show that our approach remains effective across the entire spectrum of $r$, i.e. both when the number of clean samples is relatively small, and when it is relatively large.

\subsection{Varying cost scales}\label{apx:vaying_cost}
\begin{figure}[t!]
    \centering
    \includegraphics[width=0.6\textwidth]{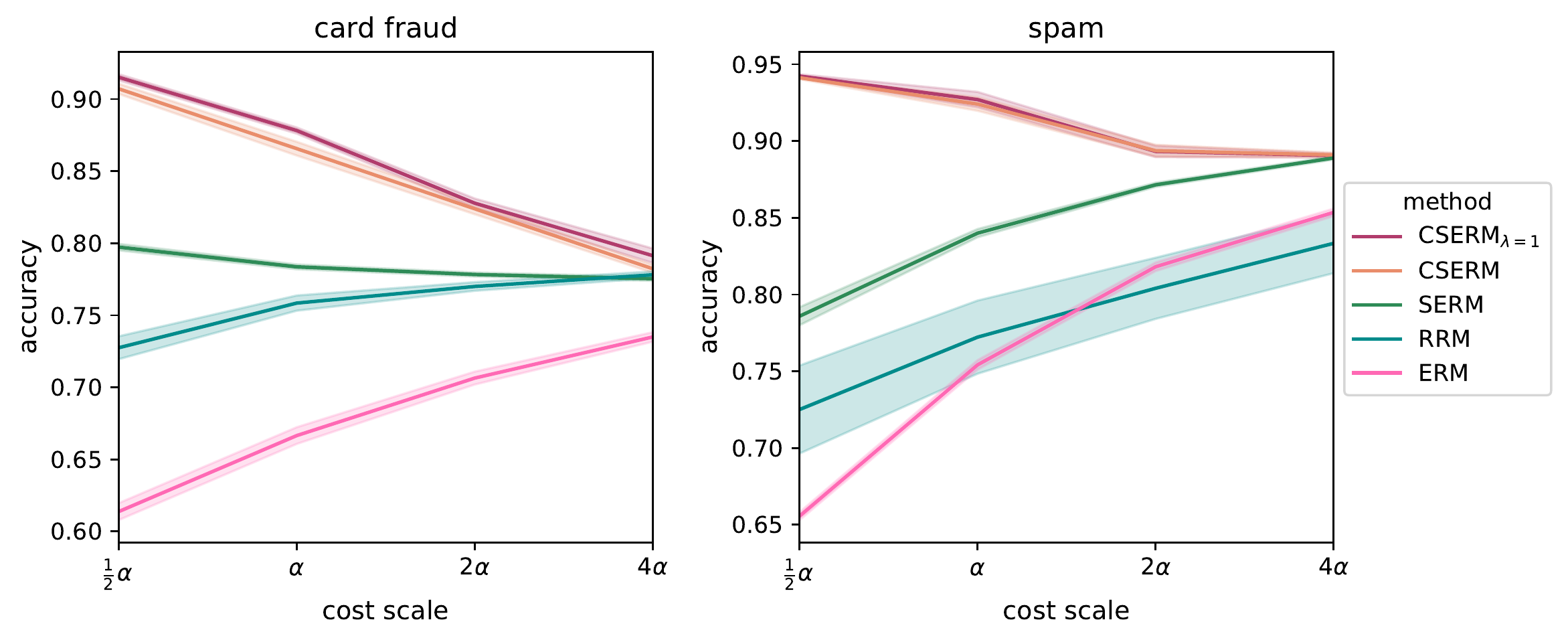}
    \caption{
    Accuracy across different cost scales.
    \nir{maybe also plot \%move? (eg in a plot below)}
    }
    \label{fig:real_exp_across_alphas}
\end{figure}

In this section we report results for all methods and for multiple cost scales $\alpha$.
Our results in the main paper (Table~\ref{tbl:real_experiments} in Sec.~\ref{sec:exp_real}) show performance for $\alpha$ chosen such that $\sim 50\%$ of points move (per dataset):
in \texttt{card fraud} we set $\alpha=1$,
and in \texttt{spam} we set $\alpha=40$.
Here we show results for other cost scales, including
$\frac{1}{2}\alpha$, $\alpha$, $2\alpha$, and $4\alpha$.
Figure \ref{fig:real_exp_across_alphas} plots performance as a function of $\alpha$.
As can be seen, in each dataset the relations between the accuracies of the baseline remain similar across different cost scales, but as the cost scale decreases, there is less movement, and the absolute gap between the methods decreases as well.
The next pages include tables reporting full results for all considered cost scales, first for \texttt{card fraud}, and then for \texttt{spam}.

\vspace{7cm}

\begin{tabular}{lrccccccc}
  &   & \multicolumn{7}{c}{\boldmath{}\textbf{\texttt{card fraud}, cost scale $\frac{1}{2}\alpha$}\unboldmath{}} \\
\cmidrule{3-9}  &   & accuracy & perceived & \%improve & \%move & \%neg${\mapsto}$pos & \%pos${\mapsto}$neg & welfare \\
\cmidrule{1-1}\cmidrule{3-9}\CSERMreg{$\lambda=1$} &   & 91.5\tiny{\,$\pm0.2$} & 95.2 & 15.3 & 59.9 & 16.9 & 1.5 & -0.67 \\
\CSERMreg{$\lambda=0.1$} &   & 91.2\tiny{\,$\pm0.3$} & 94.9 & 15.0 & 59.6 & 16.4 & 1.5 & -0.7 \\
\CSERM &   & 90.7\tiny{\,$\pm0.4$} & 95.0 & 14.4 & 59.7 & 16.0 & 1.6 & -0.57 \\
\SERM &   & 79.7\tiny{\,$\pm0.2$} & 77.5 & 2.2 & 55.5 & 3.4 & 1.2 & -0.24 \\
\RRM &   & 72.7\tiny{\,$\pm0.8$} & 73.5 & 0.7 & 24.6 & 0.9 & 0.2 & 0.05 \\
\RRMlet &   & 69.6\tiny{\,$\pm0.2$} & 77.7 & 0.3 & 16.1 & 0.3 & 0.0 & 0.2 \\
\RRMc &   & 60.8\tiny{\,$\pm0.3$} & 74.8 & 0.4 & 22.8 & 0.5 & 0.0 & 0.4 \\
\ERM &   & 61.4\tiny{\,$\pm0.6$} & 77.5 & 0.6 & 26.7 & 0.6 & 0.0 & 0.30 \\
\cmidrule{1-1}\cmidrule{3-9}\CSERMoracle &   & 89.8\tiny{\,$\pm0.2$} & 94.2 & 13.0 & 58.5 & 14.4 & 1.4 & -0.72 \\
\bench &   & 77.5\tiny{\,$\pm0.2$} & - & - & - & - & - & - \\
\end{tabular}%

\begin{tabular}{lrccccccc}
  &   & \multicolumn{7}{c}{\boldmath{}\textbf{\texttt{card fraud}, cost scale $\alpha$}\unboldmath{}} \\
\cmidrule{3-9}  &   & accuracy & perceived & \%improve & \%move & \%neg${\mapsto}$pos & \%pos${\mapsto}$neg & welfare \\
\cmidrule{1-1}\cmidrule{3-9}\CSERMreg{$\lambda=1$} &   & 87.8\tiny{\,$\pm0.2$} & 93.5 & 12.2 & 60.1 & 13.8 & 1.7 & -0.65 \\
\CSERMreg{$\lambda=0.1$} &   & 87.7\tiny{\,$\pm0.2$} & 93.6 & 11.4 & 58.9 & 13.0 & 1.6 & -0.5 \\
\CSERM &   & 86.6\tiny{\,$\pm0.5$} & 93.4 & 10.2 & 58.9 & 11.8 & 1.5 & -0.48 \\
\SERM &   & 78.4\tiny{\,$\pm0.2$} & 77.5 & 0.8 & 45.9 & 1.6 & 0.7 & -0.16 \\
\RRM &   & 75.8\tiny{\,$\pm0.5$} & 70.5 & 0.5 & 24.7 & 0.7 & 0.2 & -0.06 \\
\RRMlet &   & 71.6\tiny{\,$\pm0.2$} & 77.6 & 0.2 & 12.5 & 0.2 & 0.0 & 0.2 \\
\RRMc &   & 63.6\tiny{\,$\pm0.3$} & 74.7 & 0.3 & 18.8 & 0.3 & 0.0 & 0.4 \\
\ERM &   & 66.7\tiny{\,$\pm0.6$} & 77.5 & 0.3 & 19.8 & 0.4 & 0.0 & 0.25 \\
\cmidrule{1-1}\cmidrule{3-9}\CSERMoracle &   & 87.0\tiny{\,$\pm0.2$} & 93.3 & 10.1 & 57.9 & 11.8 & 1.6 & -0.60 \\
\bench &   & 77.5\tiny{\,$\pm0.2$} & - & - & - & - & - & - \\
\end{tabular}%

\begin{tabular}{lrccccccc}
  &   & \multicolumn{7}{c}{\boldmath{}\textbf{\texttt{card fraud}, cost scale $2\alpha$}\unboldmath{}} \\
\cmidrule{3-9}  &   & accuracy & perceived & \%improve & \%move & \%neg${\mapsto}$pos & \%pos${\mapsto}$neg & welfare \\
\cmidrule{1-1}\cmidrule{3-9}\CSERMreg{$\lambda=1$} &   & 82.8\tiny{\,$\pm0.3$} & 92.4 & 6.3 & 57.8 & 7.5 & 1.2 & -0.58 \\
\CSERMreg{$\lambda=0.1$} &   & 82.3\tiny{\,$\pm0.5$} & 92.3 & 6.5 & 58.8 & 7.7 & 1.2 & -0.7 \\
\CSERM &   & 82.4\tiny{\,$\pm0.4$} & 92.8 & 5.8 & 57.9 & 7.0 & 1.2 & -0.52 \\
\SERM &   & 77.8\tiny{\,$\pm0.1$} & 77.6 & 0.3 & 22.7 & 0.7 & 0.4 & -0.05 \\
\RRM &   & 77.0\tiny{\,$\pm0.3$} & 69.9 & 0.1 & 22.9 & 0.3 & 0.3 & -0.11 \\
\RRMlet &   & 73.6\tiny{\,$\pm0.1$} & 77.5 & 0.1 & 8.7 & 0.1 & 0.0 & 0.2 \\
\RRMc &   & 64.9\tiny{\,$\pm0.3$} & 74.2 & 0.2 & 15.9 & 0.2 & 0.0 & 0.4 \\
\ERM &   & 70.6\tiny{\,$\pm0.4$} & 77.5 & 0.1 & 13.5 & 0.2 & 0.0 & 0.22 \\
\cmidrule{1-1}\cmidrule{3-9}\CSERMoracle &   & 83.6\tiny{\,$\pm0.1$} & 91.4 & 6.7 & 56.9 & 8.0 & 1.3 & -0.53 \\
\bench &   & 77.5\tiny{\,$\pm0.2$} & - & - & - & - & - & - \\
\end{tabular}%

\begin{tabular}{lrccccccc}
  &   & \multicolumn{7}{c}{\boldmath{}\textbf{\texttt{card fraud}, cost scale $4\alpha$}\unboldmath{}} \\
\cmidrule{3-9}  &   & accuracy & perceived & \%improve & \%move & \%neg${\mapsto}$pos & \%pos${\mapsto}$neg & welfare \\
\cmidrule{1-1}\cmidrule{3-9}\CSERMreg{$\lambda=1$} &   & 79.1\tiny{\,$\pm0.5$} & 85.9 & 2.5 & 35.4 & 3.1 & 0.6 & -0.30 \\
\CSERMreg{$\lambda=0.1$} &   & 79.2\tiny{\,$\pm0.4$} & 88.2 & 2.3 & 42.0 & 3.0 & 0.6 & -0.4 \\
\CSERM &   & 78.2\tiny{\,$\pm0.3$} & 85.3 & 0.8 & 29.1 & 1.1 & 0.4 & -0.14 \\
\SERM &   & 77.5\tiny{\,$\pm0.1$} & 77.5 & 0.0 & 7.3 & 0.2 & 0.2 & 0.06 \\
\RRM &   & 77.8\tiny{\,$\pm0.2$} & 70.9 & 0.4 & 18.8 & 0.5 & 0.1 & -0.08 \\
\RRMlet &   & 74.8\tiny{\,$\pm0.2$} & 77.7 & 0.0 & 6.6 & 0.1 & 0.0 & 0.2 \\
\RRMc &   & 66.6\tiny{\,$\pm0.3$} & 74.0 & 0.2 & 12.9 & 0.2 & 0.0 & 0.3 \\
\ERM &   & 73.5\tiny{\,$\pm0.3$} & 77.5 & 0.1 & 8.7 & 0.1 & 0.0 & 0.20 \\
\cmidrule{1-1}\cmidrule{3-9}\CSERMoracle &   & 79.6\tiny{\,$\pm0.5$} & 85.6 & 2.3 & 38.4 & 2.9 & 0.6 & -0.37 \\
\bench &   & 77.5\tiny{\,$\pm0.2$} & - & - & - & - & - & - \\
\end{tabular}%

\begin{tabular}{lrccccccc}
  &   & \multicolumn{7}{c}{\boldmath{}\textbf{\texttt{spam}, cost scale $\frac{1}{2}\alpha$}\unboldmath{}} \\
\cmidrule{3-9}  &   & accuracy & perceived & \%improve & \%move & \%neg${\mapsto}$pos & \%pos${\mapsto}$neg & welfare \\
\cmidrule{1-1}\cmidrule{3-9}\CSERMreg{$\lambda=1$} &   & 94.2\tiny{\,$\pm0.1$} & 97.6 & 6.4 & 57.0 & 6.4 & 0.0 & -0.27 \\
\CSERMreg{$\lambda=0.1$} &   & 94.0\tiny{\,$\pm0.2$} & 97.9 & 5.8 & 53.9 & 5.8 & 0.0 & -0.3 \\
\CSERM &   & 94.1\tiny{\,$\pm0.1$} & 97.6 & 5.8 & 52.8 & 5.8 & 0.0 & -0.25 \\
\SERM &   & 78.6\tiny{\,$\pm0.6$} & 91.2 & -12.6 & 56.0 & 0.0 & 12.6 & -0.36 \\
\RRM &   & 72.5\tiny{\,$\pm2.9$} & 76.9 & 0.4 & 36.1 & 1.3 & 0.9 & 0.13 \\
\RRMlet &   & 74.0\tiny{\,$\pm0.8$} & 90.9 & 0.1 & 21.0 & 0.3 & 0.2 & 0.2 \\
\RRMc &   & 74.9\tiny{\,$\pm0.6$} & 87.1 & -1.2 & 21.6 & 0.2 & 1.4 & 0.2 \\
\ERM &   & 65.6\tiny{\,$\pm0.3$} & 91.2 & 0.4 & 27.0 & 0.4 & 0.0 & 0.26 \\
\cmidrule{1-1}\cmidrule{3-9}\CSERMoracle &   & 94.4\tiny{\,$\pm0.1$} & 94.4 & 6.1 & 58.6 & 6.1 & 0.0 & -0.31 \\
\bench &   & 91.2\tiny{\,$\pm0.1$} & - & - & - & - & - & - \\
\end{tabular}%

\begin{tabular}{lrccccccc}
  &   & \multicolumn{7}{c}{\boldmath{}\textbf{\texttt{spam}, cost scale $\alpha$}\unboldmath{}} \\
\cmidrule{3-9}  &   & accuracy & perceived & \%improve & \%move & \%neg${\mapsto}$pos & \%pos${\mapsto}$neg & welfare \\
\cmidrule{1-1}\cmidrule{3-9}\CSERMreg{$\lambda=1$} &   & 92.7\tiny{\,$\pm0.5$} & 97.0 & 3.1 & 37.5 & 3.2 & 0.1 & -0.23 \\
\CSERMreg{$\lambda=0.1$} &   & 92.6\tiny{\,$\pm0.4$} & 97.3 & 2.9 & 37.2 & 2.9 & 0.0 & -0.1 \\
\CSERM &   & 92.4\tiny{\,$\pm0.4$} & 97.1 & 2.4 & 36.7 & 2.5 & 0.1 & -0.21 \\
\SERM &   & 84.0\tiny{\,$\pm0.3$} & 91.2 & -7.2 & 41.3 & 0.0 & 7.2 & -0.17 \\
\RRM &   & 77.2\tiny{\,$\pm2.4$} & 76.6 & -2.1 & 30.5 & 0.6 & 2.8 & 0.07 \\
\RRMlet &   & 78.7\tiny{\,$\pm0.6$} & 91.2 & 0.1 & 14.8 & 0.3 & 0.1 & 0.2 \\
\RRMc &   & 79.1\tiny{\,$\pm0.6$} & 87.4 & -0.8 & 15.8 & 0.2 & 1.1 & 0.2 \\
\ERM &   & 75.4\tiny{\,$\pm0.3$} & 91.2 & 0.4 & 17.1 & 0.4 & 0.0 & 0.23 \\
\cmidrule{1-1}\cmidrule{3-9}\CSERMoracle &   & 93.5\tiny{\,$\pm0.1$} & 93.5 & 4.5 & 41.7 & 4.5 & 0.0 & -0.17 \\
\bench &   & 91.2\tiny{\,$\pm0.1$} & - & - & - & - & - & - \\
\end{tabular}%

\begin{tabular}{lrccccccc}
  &   & \multicolumn{7}{c}{\boldmath{}\textbf{\texttt{spam}, cost scale $2\alpha$}\unboldmath{}} \\
\cmidrule{3-9}  &   & accuracy & perceived & \%improve & \%move & \%neg${\mapsto}$pos & \%pos${\mapsto}$neg & welfare \\
\cmidrule{1-1}\cmidrule{3-9}\CSERMreg{$\lambda=1$} &   & 89.3\tiny{\,$\pm0.4$} & 94.3 & -1.3 & 21.6 & 0.2 & 1.5 & -0.03 \\
\CSERMreg{$\lambda=0.1$} &   & 88.9\tiny{\,$\pm0.3$} & 95.8 & -1.8 & 22.4 & 0.1 & 1.9 & 0.0 \\
\CSERM &   & 89.4\tiny{\,$\pm0.4$} & 95.1 & -1.4 & 22.0 & 0.2 & 1.5 & -0.05 \\
\SERM &   & 87.1\tiny{\,$\pm0.1$} & 91.2 & -4.1 & 25.7 & 0.0 & 4.1 & -0.07 \\
\RRM &   & 80.4\tiny{\,$\pm2.0$} & 79.0 & -2.3 & 22.1 & 0.3 & 2.6 & 0.00 \\
\RRMlet &   & 82.6\tiny{\,$\pm0.6$} & 90.9 & 0.1 & 9.7 & 0.2 & 0.1 & 0.2 \\
\RRMc &   & 81.7\tiny{\,$\pm0.4$} & 87.5 & -0.4 & 11.5 & 0.2 & 0.6 & 0.2 \\
\ERM &   & 81.8\tiny{\,$\pm0.3$} & 91.2 & 0.2 & 10.5 & 0.2 & 0.0 & 0.22 \\
\cmidrule{1-1}\cmidrule{3-9}\CSERMoracle &   & 89.0\tiny{\,$\pm0.3$} & 89.0 & -1.9 & 23.2 & 0.0 & 1.9 & -0.04 \\
\bench &   & 91.2\tiny{\,$\pm0.1$} & - & - & - & - & - & - \\
\end{tabular}%

\begin{tabular}{lrccccccc}
  &   & \multicolumn{7}{c}{\boldmath{}\textbf{\texttt{spam}, cost scale $4\alpha$}\unboldmath{}} \\
\cmidrule{3-9}  &   & accuracy & perceived & \%improve & \%move & \%neg${\mapsto}$pos & \%pos${\mapsto}$neg & welfare \\
\cmidrule{1-1}\cmidrule{3-9}\CSERMreg{$\lambda=1$} &   & 89.1\tiny{\,$\pm0.1$} & 93.5 & -1.6 & 11.8 & 0.0 & 1.6 & 0.06 \\
\CSERMreg{$\lambda=0.1$} &   & 89.0\tiny{\,$\pm0.2$} & 93.2 & -1.6 & 11.9 & 0.0 & 1.6 & 0.1 \\
\CSERM &   & 89.1\tiny{\,$\pm0.2$} & 93.4 & -1.7 & 12.4 & 0.0 & 1.7 & 0.07 \\
\SERM &   & 88.9\tiny{\,$\pm0.1$} & 91.2 & -2.3 & 15.4 & 0.0 & 2.3 & 0.01 \\
\RRM &   & 83.3\tiny{\,$\pm1.9$} & 81.2 & -0.8 & 16.3 & 0.2 & 1.0 & 0.02 \\
\RRMlet &   & 84.8\tiny{\,$\pm0.4$} & 91.0 & 0.0 & 7.2 & 0.1 & 0.1 & 0.2 \\
\RRMc &   & 83.6\tiny{\,$\pm0.2$} & 87.4 & -0.2 & 8.3 & 0.1 & 0.3 & 0.2 \\
\ERM &   & 85.4\tiny{\,$\pm0.2$} & 91.2 & 0.0 & 6.8 & 0.0 & 0.0 & 0.20 \\
\cmidrule{1-1}\cmidrule{3-9}\CSERMoracle &   & 89.1\tiny{\,$\pm0.1$} & 89.1 & -1.9 & 13.2 & 0.0 & 1.9 & 0.05 \\
\bench &   & 91.2\tiny{\,$\pm0.1$} & - & - & - & - & - & - \\
\end{tabular}%


\end{document}